\documentclass[preprint,12pt]{elsarticle}
\usepackage{amsmath,amssymb,graphicx,multirow}
\usepackage[ruled,vlined]{algorithm2e}
\usepackage{array}
\usepackage{wrapfig}
\usepackage{subcaption}

\usepackage[utf8]{inputenc} % allow utf-8 input
\usepackage[T1]{fontenc}    % use 8-bit T1 fonts
\usepackage{hyperref}       % hyperlinks
\usepackage{url}            % simple URL typesetting
\usepackage{booktabs}       % professional-quality tables
\usepackage{amsfonts}       % blackboard math symbols
\usepackage{nicefrac}       % compact symbols for 1/2, etc.
\usepackage{microtype}      % microtypography
\usepackage{xcolor}         % colors

\journal{Neurocomputing}

\begin{document}

\begin{frontmatter}

\title{Federated Unsupervised Semantic Segmentation}

\author{Evangelos Charalampakis} %% Author name
\author{Vasileios Mygdalis}
\author{Ioannis Pitas}
\address{ Department of Informatics,
Aristotle University of Thessaloniki,
Thessaloniki, Greece \\
\{charevan, mygdalisv, pitas\}@csd.auth.gr}
% \affiliation{organization={School of Informatics, Aristotle University of Thessaloniki},%Department and Organization
%             city={Thessaloniki},
%             country={Greece}}

% \affiliation{organization={School of Informatics, Aristotle University of Thessaloniki},%Department and Organization
%             city={Thessaloniki},
%             country={Greece}}

%% Abstract
\begin{abstract}
This work explores the application of Federated Learning (FL) to Unsupervised Semantic image Segmentation (USS). 
Recent USS methods extract pixel-level features using frozen visual foundation models and refine them through self-supervised objectives that encourage semantic grouping. These features are then grouped to semantic clusters to produce segmentation masks.
Extending these ideas to federated settings requires feature representation and cluster centroid alignment across distributed clients, an inherently difficult task under heterogeneous data distributions in the absence of supervision.
To address this, we propose \textbf{FUSS} (\textbf{F}ederated \textbf{U}nsupervised image \textbf{S}emantic \textbf{S}egmentation) which is, to our knowledge, the first framework to enable fully decentralized, label-free semantic
segmentation training. FUSS introduces novel federation
strategies that promote global consistency in feature and prototype space, jointly optimizing local segmentation heads and shared semantic centroids. Experiments on both benchmark and real-world datasets, including binary and multi-class segmentation tasks, show that FUSS consistently outperforms local-only client trainings as well as extensions of classical FL algorithms under varying client data distributions. To fully support reproducibility, the source code, data partitioning scripts, and implementation details are publicly available at: \url{https://github.com/evanchar/FUSS}
\end{abstract}


% \begin{highlights}
% \item Problem definition of Federated Unsupervised Semantic Segmentation (FUSS)
% \item Various federated aggregation strategies are examined
% \item FedCC: Novel prototype alignment strategies for heterogeneous clients
% \item Experimental results show improved performance over federated baselines.
% \end{highlights}

%% Keywords
\begin{keyword}
Federated Learning \sep Image Segmentation \sep Unsupervised Learning \sep Prototype Alignment \sep FedCC

%% or \MSC[2008] code \sep code (2000 is the defaults)

\end{keyword}

\end{frontmatter}

\section{Introduction}\label{sec:intro}
Semantic image segmentation is a fundamental computer vision task, with the objective of assigning image pixels to pre-defined class labels. Given its central role in scene understanding, semantic segmentation serves as a core component of visual perception systems deployed in safety-critical applications like autonomous driving \cite{tang2022perception}, industrial inspection \cite{psarras2024unified}, or medical diagnostics \cite{chen2024think, rezaei2021review}. 
Unsupervised Semantic Segmentation (USS) has emerged as a family of methods that do not require pixel-wise annotations, which are notoriously expensive. While the absence of manual annotations makes it infeasible to directly assign semantic labels to image regions, USS methods aim to discover coherent visual groupings whose emergent categories exhibit strong alignment with human-perceived semantic concepts.
Recent USS approaches employ pretrained Convolutional Neural Networks (CNNs) \cite{he2016deep} or Vision Foundation Models (VFMs) \cite{caron2021emerging, dosovitskiy2020image} to extract pixel-wise feature embeddings. These representations are fed into learnable segmentation heads that are fine-tuned through self-supervised learning objectives such as contrastive losses \cite{zhang2024dynaseg, hahn2024boosting}, energy minimization \cite{hamilton2022unsupervised}, or eigenvector  alignment \cite{kim2024eagle}. Unsupervised segmentation is ultimately achieved by simply clustering the learned representations.

In safety-critical domains, strict privacy constraints on data storage and sharing introduce significant challenges in effectively training robust and generalizable DNN-based image segmentation models. For instance, hospitals routinely collect large volumes of unlabeled radiology or pathology scans but are unable to share them across institutions due to privacy regulations such as HIPAA \cite{gostin2009beyond} or GDPR \cite{voigt2017eu}. This restriction can lead to biased or under-trained segmentation models. Similarly, industrial facilities may require semantic segmentation models for pipeline and machinery visual inspection \cite{psarras2024unified}, yet they face both annotation scarcity and limitations on transmitting proprietary data, which may lead to undetected defects and increased operational risk. In the case of autonomous vehicles, edge devices must adapt segmentation models to local video streams without access to annotations \cite{sahu2022clustering} and under strict data-sharing constraints, where failure to do so can induce bias or silent model degradation, with potentially severe safety consequences.

To this end, Federated Learning (FL) \cite{mcmahan2017communication,caldarola2022improving, gong2021ensemble, fantauzzo2022feddrive, vyas2023federated} has emerged as a promising paradigm that enables collaborative model training with decentralized data. Perhaps the most recognizable FL method is FedAvg \cite{mcmahan2017communication}, a method that simply aggregates the weights between federation clients. The simplicity and elegance of FedAvg render it applicable to almost every supervised \cite{samad2025fedgclrec} or unsupervised training \citep{lubana2022orchestra,liao2024rethinking} scenario, making FedAvg the standard benchmark in most FL applications.

% FUSS is motivated by domains where both data sharing and annotation efforts are prohibitively costly or restricted, conditions that are becoming increasingly prevalent. 

In semantic image segmentation, federated learning (FL) has thus far been investigated primarily under supervised training regimes \cite{miao2023fedseg, fantauzzo2022feddrive}, or within domain adaptation frameworks that rely on large annotated datasets for source-domain pretraining \cite{yao2022federated, shenaj2023learning}. 
% Old key-challende
% A key challenge in this context arises from the combined effect of data heterogeneity across clients and substantial within-class variance caused by fine-grained visual differences (i.e., pixels belonging to semantically similar objects may exhibit significant appearance variation). 
A key challenge in this context arises from two distinct forms of data heterogeneity. First, domain shift (feature distribution skew) that occurs as a natural characteristic of real-world deployments. For example, distinct visual sensors may capture different backgrounds or lighting conditions. Furthermore, the same semantic object may exhibit heavy appearance variation in different locations (e.g., a pristine pipeline in a modern refinery versus a damaged pipeline in an unsupported industrial site). Second, label distribution skew (class imbalance) that forces clients to discover semantic prototypes from disjoint class subsets. In unsupervised settings with a fixed ``class discovery'' budget, this form of heterogeneity can be particularly challenging, as the absence of specific classes, may force local models to assign ``empty'' prototype slots to random features or noise, disrupting alignment across federated clients.

Supervised objectives, as demonstrated in \cite{miao2023fedseg}, can partially mitigate this issue by aligning representations of similar objects toward consistent target values across clients. However, these mechanisms are inherently inapplicable in the Federated Unsupervised Semantic Segmentation (FUSS) setting, where no supervised signal is available at any stage of the segmentation pipeline. Consequently, learning coherent and transferable semantic representations under these constraints remains an open and largely unexplored research problem.

To bridge this gap, we propose \textbf{F}ederated \textbf{U}nsupervised image \textbf{S}emantic \textbf{S}egmentation (\textbf{FUSS}), a novel framework for collaborative unsupervised training of image segmentation DNN models in decentralized settings. FUSS addresses both data annotation scarcity and privacy constraints by
enabling clients to jointly learn DNN segmentation heads and class prototypes using unsupervised objectives and communication-efficient parameter exchanges.

Across these domains, a common challenge emerges: When an entity requires an AI system for high-risk, safety-critical tasks but lacks adequate, unbiased, or annotated data (due to privacy or time constraints), centralized training cannot be trusted to produce reliable results. In such scenarios, where centralized annotation or data sharing is infeasible, \textbf{federated unsupervised training is not just practical, it may be the only viable solution}.

The key contributions of this paper are summarized as follows:
\begin{itemize}    
    \item 
    We formally define the \textbf{FUSS framework}, which combines unsupervised representation learning with semantic prototype alignment across clients. FUSS enables efficient and stable federated image segmentation training without any reliance on annotated image data, a characteristic that allows seamless applicability in heavily privacy-constrained data settings.

    \item 
    We develop a novel aggregation strategy, namely \textbf{Federated Centroid Clustering (FedCC)}, tailored to federated unsupervised segmentation. FedCC supports joint optimization of client-specific segmentation heads and global alignment of semantic class prototypes, enabling effective end-to-end training in decentralized settings, even with highly heterogeneous clients.

    \item  We establish the first comprehensive benchmarks for FUSS on the Cityscapes and CocoStuff datasets and validate our methods on a real-world industrial pipeline segmentation dataset. 
\end{itemize}
Through extensive experimentation, we demonstrate that USS, combined with FedCC, consistently outperforms local DNN training and extensions of traditional FL aggregation strategies in fully unsupervised federated scenarios, while preserving scalability and data privacy.

The remainder of this paper is structured as follows. Section~\ref{sec:Related} provides an overview of related work in unsupervised semantic image segmentation and federated learning. Section~\ref{sec:Preliminaries} introduces USS unified notation. Section~\ref{sec:FUSS} presents the proposed FUSS framework, detailing our methodology, architectural choices and aggregation strategies. Section~\ref{sec:experiments} describes the experimental setup, including dataset partitions and federated configurations, and reports quantitative results along with key observations. Finally, Section~\ref{sec:conclusion} concludes the paper and outlines directions for future research.

\section{Related Work}\label{sec:Related}
\subsection{Unsupervised Semantic Segmentation}
Early Unsupervised Semantic Segmentation (USS) approaches were not DNN-based \citep{pitas2000digital}. Instead, they relied on low-level hand-crafted image features, such as color histograms, texture, and intensity gradients, to cluster pixels into coherent groups. While effective for simpler scenes, these methods often struggled with the complex semantic variability found in unrestricted real-world environments.

To mitigate this, earlier deep learning approaches introduced end-to-end unsupervised objectives to train Convolutional Neural Networks (CNNs). For instance, W-Net \citep{xia2017w} concatenated two U-Net architectures into an autoencoder, jointly optimizing a soft normalized cut loss for segmentation and a reconstruction loss for data fidelity. DeepCluster \citep{caron2018deep} proposed an iterative framework where features from a CNN were clustered via k-means to generate pseudo-labels, which then supervised the network's weight updates, effectively bootstrapping semantic representations. 

More recently, PiCIE \citep{cho2021picie} advanced this direction by employing invariance to photometric transformations and equivariance to geometric shifts as inductive biases, enabling the clustering of pixels into semantically coherent groups even in non-object-centric datasets. However, despite these advancements, such methods often required extensive training from scratch and struggled to capture the rich, open-world semantic granularity offered by modern Vision Foundation Models.

With the emergence of self-supervised representation learning \citep{gui2024survey}, USS saw significant progress. STEGO (Self-supervised Transformer with Energy-based Graph Optimization) \citep{hamilton2022unsupervised} was a key breakthrough,
which used self-supervised Vision Transformers \citep{caron2021emerging} pre-trained on contrastive objectives to generate semantically structured feature maps, where pixels belonging to similar objects or regions within an image are mapped to proximate locations in the embedding space.

Building on this, EAGLE (Eigen Aggregation Learning for Object-Centric Unsupervised Semantic Segmentation) \citep{kim2024eagle} introduced a spectral image segmentation approach that utilizes structural cues derived from the image color affinity graph to produce more object-centric and boundary-aware image segmentation. 
PriMaPs \citep{hahn2024boosting} introduced a novel paradigm that decomposes images into semantically meaningful ``principal mask proposals'' based on feature representations. By fitting class prototypes to these proposals via a stochastic expectation-maximization algorithm (PriMaPs-EM), 
the method effectively boosts unsupervised segmentation performance. 

While these approaches offer refinements in specific scenarios, they often introduce additional computational overhead for incremental gains. STEGO remains a cornerstone of the field, providing highly competitive
% —and in some regimes, superior—
performance with significantly lower complexity. This balance of performance and efficiency motivates our adoption of STEGO as the core segmentation engine for our federated framework.

Although distinct in their objectives, the adjacent fields of Unsupervised Salient Object Detection (USOD) and Instance Segmentation have recently proposed mechanisms highly transferable to the USS context. In the domain of USOD, \citet{liu2024deep} introduced a Belief Capsule Network (BCNet) that captures object-part hierarchies, enabling the generation of refined pseudo-masks with enhanced object integrity. Moreover, for instance segmentation in data-scarce regimes, \citet{zhang2025unsupervised} leveraged Vision-Language Models to generate semantically rich pseudo-masks, which subsequently serve as prompts for unsupervised pre-training of Query-based End-to-End Image Segmentation models (QEIS). These innovations introduce strong inductive biases through hierarchical grouping and language-driven guidance, that inspire directions for overcoming the granularity and initialization challenges inherent in unsupervised scene understanding.

Despite architectural and optimization differences, SOTA USS methods share a common foundation that also motivates our approach: the use of strong self-supervised DNN backbones for feature extraction, followed by unsupervised optimization techniques to uncover semantically meaningful groupings in the data. 
% In the following subsection we introduce the necessary notation.

\subsection{Federated Learning}
Federated Learning (FL) \citep{mcmahan2017communication} has emerged as the standard for collaborative model training under privacy constraints, enabling distributed optimization without centralizing raw data. In the specific domain of semantic segmentation, existing FL frameworks have primarily focused on supervised settings, where clients are assumed to possess pixel-perfect ground-truth labels, but nontheless appear to be statistically heterogeneous, and attribute inherent in distributed environments.

To address this challenge, methods like FedDrive \citep{fantauzzo2022feddrive} introduced effective benchmarking strategies for domain generalization, utilizing style transfer to bridge the domain shift across autonomous driving datasets. Similarly, FedSeg \citep{miao2023fedseg} tackled high label distribution skew by correcting client local optimization with a modified cross-entropy loss and employing pixel-level contrastive learning to align local embeddings with a global semantic space.

While effective, these methods remain fundamentally constrained by their reliance on extensive annotated data, which is often unavailable or prohibitively expensive to acquire in real-world edge scenarios.

Parallel efforts have explored Unsupervised Federated Learning (UFL) to eliminate the dependency on labels. Approaches such as Orchestra \citep{lubana2022orchestra} and FedU \citep{zhuang2021collaborative} have demonstrated that global clustering mechanisms and divergence-aware aggregation can yield robust representations even from non-IID data. \citet{liao2024rethinking} further advanced this by addressing representation collapse through a flexible uniform regularizer that preserves uniformity in the embedding space. However, these works predominantly target image-level classification tasks. They lack the dense, spatial granularity required to enforce semantic consistency at the pixel level, leaving the challenge of Federated Unsupervised Semantic Segmentation largely unaddressed.

To the best of our knowledge, FUSS is the first framework specifically designed to bridge this gap. Unlike previous methods that optimize for global image categories, FUSS introduces a novel paradigm for jointly aligning pixel-level feature representations and global cluster centroids in a fully decentralized, label-free manner. By extending the principles of self-supervised grouping to the federated setting, our approach enables the discovery of coherent semantic regions across heterogeneous clients without sharing raw data or annotations.

\section{Preliminaries}\label{sec:Preliminaries}
Formally, given an unlabeled image dataset $ \mathcal{D} = \{ \mathbf{X}_i \in \mathbb{R}^{H\times W \times 3} \mid i = 1, \dots, N \} $, the goal of USS is to estimate a segmentation mask $ \hat{\mathbf{Y}}_i \in \{0,1\}^{H \times W \times |\mathcal{C}|} $ for each image, where $\mathcal{C}$ denotes the set of semantic regions or latent classes to be discovered in the data. In the absence of ground-truth labels, the value of $|\mathcal{C}|$ is assumed to be known and remains fixed. 
\\\\
\textbf{Feature extraction} Let $E$ denote a pretrained DNN encoder that maps an image $\mathbf{X}_i$ to a dense feature representation $\mathbf{H}_i = E(\mathbf{X}_i)$, with $\mathbf{H}_i \in \mathbb{R}^{H \times W \times D}$. Each location $(h, w)$ in $\mathbf{H}_i$ corresponds to a $D$-dimensional embedding capturing local semantic context.

To enable effective clustering, the encoder must generate feature representations in which pixels belonging to the same semantic object exhibit high similarity, while pixels from different objects remain clearly separable. We formalize this by defining a general similarity function:
\begin{equation}
    s(\mathbf{H}^{(h,w)}, \mathbf{H}^{(m,n)}) = \phi\left( \psi(\mathbf{H}^{(h,w)}, \mathbf{H}^{(m,n)}) \right),
\end{equation}
where $\psi: \mathbb{R}^D \times \mathbb{R}^D \rightarrow \mathbb{R}$ denotes a base similarity metric, such as cosine similarity, and $\phi: \mathbb{R} \rightarrow \mathbb{R}$ represents an optional normalization or scaling function. Tuple $(m,n)$ represents the spatial coordinates (row and column indices) of a second pixel in the feature map $\mathbf{H}$, distinct from the reference pixel at $(h,w)$. This general formulation permits the adoption of various similarity functions within the USS framework. Larger values of $s$ correspond to stronger semantic affinity between pixel-level representations, whereas values closer to the lower bound of the function indicate greater dissimilarity between feature representations. Such a formulation allows for effective clustering-based discovery of coherent object groupings.

To guide the encoder $E$ towards learning semantically coherent pixel-wise representations, contrastive objectives are typically employed \cite{caron2021emerging}. The resulting loss functions promote intra-object compactness by increasing the similarity among perceptually related pixels, while simultaneously encouraging inter-object separability by reducing similarity between unrelated regions.

We formalize this principle with a general contrastive loss template defined over feature similarities:

\begin{equation}\label{eq:contrastive_loss}
% \scalebox{0.85}{$
\mathcal{L}_{\text{contrast}} = \sum_{(p, q) \in \mathcal{P}} \left[ - s(\mathbf{H}^{(p)}, \mathbf{H}^{(q)}) + \frac{1}{|\mathcal{N}(p)|} \sum_{r \in \mathcal{N}(p)} s(\mathbf{H}^{(p)}, \mathbf{H}^{(r)}) \right],
% $}
\end{equation}
where $\mathcal{P}$ denotes the set of positive pixel pairs in the feature map $\mathbf{H}$ and $ \mathcal{N}(p)$ the set of negatives for anchor $p$. This formulation encourages perceptually related features to be close in the embedding space, while distancing unrelated ones. Once such discriminative feature representations are obtained, the next step involves grouping them into semantically consistent regions via clustering.
\\\\
\textbf{Pixel feature clustering} 
Following feature extraction, the second stage of the USS pipeline involves grouping pixel-level feature vectors into $|\mathcal{C}|$ distinct clusters, each corresponding to a semantic region.
We define a matrix $\mathbf{M} \in \mathbb{R}^{|\mathcal{C}| \times D}$, where each row $\mathbf{m}_c$ is a cluster centroid representing a semantic prototype for the $c$-th region or object class.
% To associate pixel embeddings with a cluster, we compute a similarity matrix via the inner product:
To associate pixel embeddings with cluster prototypes, we first reshape the encoder output 
$\mathbf{H}_i \in \mathbb{R}^{H \times W \times D}$ into a matrix 
$\tilde{\mathbf{H}}_i \in \mathbb{R}^{(HW) \times D}$, where each row corresponds to the feature vector of a spatial location. 
We then compute the similarity matrix via inner product:
\begin{equation}\label{eq:similarity}
    \tilde{\mathbf{P}}_i = \tilde{\mathbf{H}}_i \mathbf{M}^\top,
\end{equation} 
where $\tilde{\mathbf{P}}_i \in \mathbb{R}^{(HW) \times |\mathcal{C}|}$ contains the unnormalized similarity scores between pixel features and centroids. 
To maintain consistency with the spatial structure of the input image, we reshape the result back into a tensor 
$\mathbf{P}_i \in \mathbb{R}^{H \times W \times |\mathcal{C}|}$.
Each element $\mathbf{P}_i^{(h,w,c)} \in \mathbb{R}$ measures the similarity between feature $\mathbf{H}_i^{(h,w)}$ and centroid $\mathbf{m}_c$.
The segmentation mask $\hat{\mathbf{Y}}_i$ is then produced by assigning each pixel to its most similar cluster:
% is obtained via a per-pixel \texttt{argmax} operation over the $C$ similarity scores:
\begin{equation}
    \hat{\mathbf{Y}}_i^{(h,w,c)} =  
    \begin{cases}
    1, & \text{if } c = \arg\max\limits_{c'} \mathbf{P}_i^{(h,w,c')}, \\
    0, & \text{otherwise}.
    \end{cases}
\end{equation}
This operation assigns each pixel to its most semantically similar cluster centroid, yielding a one-hot encoded segmentation mask.

% Accurate unsupervised segmentation requires the cluster centroids $\mathbf{m}_c$ to capture semantically meaningful structure in the feature space. However in USS, where the objects that needs to be clustered are essentially all the pixels of all the images in $\mathcal{D}$, computing all the feature representations and then producing cluster centroids with traditional clustering algorithms that utilize euclidean distance (i.e., kmeans etc), is computationally prohibitive.

The predicted masks depend critically on the quality of the centroids $\mathbf{m}_c$, which must reflect meaningful semantic structure in the embedding space. However, in USS, where all pixels across the dataset $\mathcal{D}$ are treated as clustering targets, full-dataset pixel feature extraction and traditional clustering (e.g., k-means) is computationally prohibitive.

Instead, the centroid matrix $\mathbf{M}$ is initialized randomly and incrementally optimized with each mini-batch of extracted features \cite{macqueen1967some}. The goal is for pixel representations assigned to the same centroid to be similar, while different centroids remain well-separated. This is typically achieved by minimizing intra-cluster variance and penalizing inter-cluster similarity:

\begin{equation}\label{eq:cluster_loss}
     \mathcal{L}_{\text{cluster}} = 
    \underbrace{\sum_{i,h,w} \left\| \mathbf{H}_i^{(h,w)} - \mathbf{m}_{a_{i}^{(h,w)}} \right\|^2}_{\text{intra-cluster variance}} 
    + \lambda \cdot 
    \underbrace{\sum_{c \neq c'} s(\mathbf{m}_c, \mathbf{m}_{c'})}_{\text{inter-cluster similarity}},
\end{equation}

where $a_i^{(h,w)}$ denotes the index of the centroid assigned to pixel $(h,w)$, and $\lambda > 0$ controls the trade-off between compactness and inter-cluster separation. This objective encourages compact, well-separated prototypes and enables stable pixel-to-region assignments in the absence of ground truth labels.

\section{Federated Unsupervised Semantic Segmentation (FUSS)}\label{sec:FUSS}
FUSS extends the standard Federated Learning (FL) paradigm. \cite{mcmahan2017communication, li2020federated, li2021model} to the unsupervised semantic segmentation setting, where clients must collaboratively train segmentation models without any labeled data or data sharing.

  \begin{figure*}[!htbp]
  \centering
  \includegraphics[width=\linewidth]{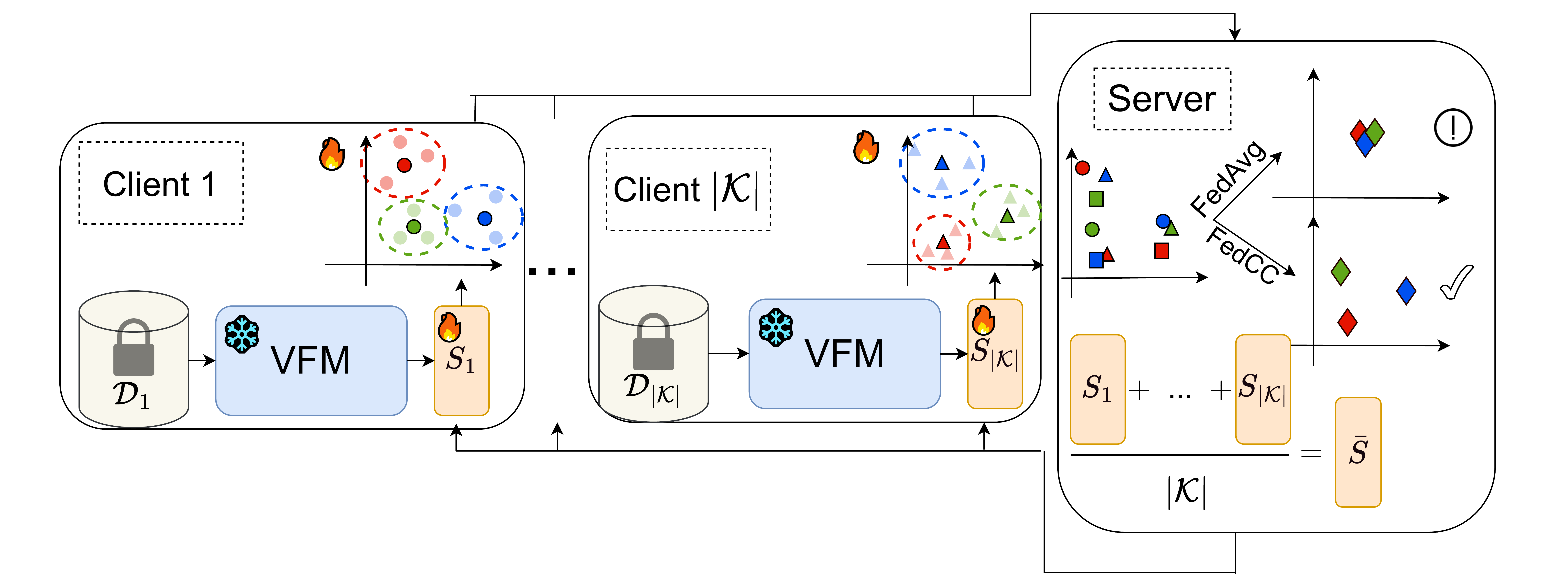}
  \caption{Overview of the proposed FUSS framework. Each client performs local unsupervised training using self-supervised objectives and periodically transmits its learned segmentation head and semantic prototypes to a central server. The server aggregates these parameters and broadcasts the updated global model back to all clients. While traditional federated strategies may result in disorganized or overlapping class prototypes due to local heterogeneity, our proposed aggregation methods explicitly promote semantic alignment and prototype separability across the federation.}
  \label{fig:fuss_overview}
\end{figure*}

Let $ \mathcal{K} $ denote the set of participating clients, each indexed by $ k $. Each client holds a private unlabeled dataset $\mathcal{D}_k = \{ \mathbf{X}_{k_i} \in \mathbb{R}^{H \times W \times 3} \mid i = 1, \dots, N_k \},$ where $ N_k $ is the number of local images. The global dataset is $ \mathcal{D} = \bigcup_{k \in \mathcal{K}} \mathcal{D}_k $, with total size $ N = \sum_{k \in \mathcal{K}} N_k $.

Each client independently trains a local encoder $ E_k $ with parameters $ \theta_{E_k} $ and maintains a local centroid matrix $ \mathbf{M}_k \in \mathbb{R}^{|\mathcal{C}| \times D} $. Local training is performed using the client’s private data and unsupervised objectives, and periodically, clients transmit their updated encoder parameters and centroids to a central server. The server aggregates these updates and broadcasts global model parameters $\bar{\theta}_E$ and aligned centroid matrix $\bar{\mathbf{M}}$ back to all clients.

Unlike traditional FL setups, FUSS must ensure consistency not only in feature extraction but also in semantic clustering. Aligning the learned prototypes $ \mathbf{M}_k $ across clients is nontrivial, as potential local data heterogeneity can induce variation in learned feature spaces and disrupt semantic alignment. To mitigate the adverse effects of data heterogeneity and class imbalance during local representation learning, FUSS leverages recent advances in vision foundation models (VFMs) such as DINO \cite{caron2021emerging}, which produce semantically meaningful pixel embeddings without supervision that coarsely align semantically similar regions even across unseen classes.

\subsection{Local training}
Following the centralized USS training of \cite{hamilton2022unsupervised}, each client in FUSS is initialized with a frozen VFM encoder, enabling efficient and stable local learning. This choice aligns with recent trends in efficient FL adaptation \cite{wang2024go}, though FUSS takes a simpler approach: by freezing the  backbone of the encoder entirely, each client performs one-time feature extraction over its local dataset, and optimizes and transmits only a segmentation head $S_{k}$ and the centroid matrix $\mathbf{M}_k$. This eliminates the need for layer scheduling or repeated backward passes, enabling streamlined training even in resource-constrained settings. 

Specifically, let $ E_\text{VFM}(\cdot, \theta_\text{VFM}) $ denote a pretrained backbone that produces intermediate feature maps:
\begin{equation}
   \mathbf{Z}_{k_i} = E_\text{VFM}(\mathbf{X}_{k_i}, \theta_\text{VFM}) \in \mathbb{R}^{H \times W \times D'}, 
\end{equation}
and $ S_k(\cdot, \theta_{S_k}) $ a lightweight, trainable projection head that maps these features to a segmentation embedding space:
\begin{equation}
\mathbf{H}_{k_i} = S_k(\mathbf{Z}_{k_i}, \theta_{S_k}) \in \mathbb{R}^{H \times W \times D}.
\end{equation}

To optimize the segmentation head, since $D << D'$ FUSS, instead of a loss in the form of Eq.~\ref{eq:contrastive_loss}, adopts a correspondence distillation loss that encourages semantic consistency across images, across features projected in different dimensions. Given feature maps $ \mathbf{Z}_{k_i} $ and $ \mathbf{Z}_{k_j} $ from two images, we define a cosine similarity tensor:
\begin{equation}
    \mathbf{A}_{k_{i,j}}^{(h,w,m,n)} := s_{\cos}(\mathbf{Z}_{k_i}^{(h,w)}, \mathbf{Z}_{k_j}^{(m,n)}),
\end{equation}
and similarly for the projected embeddings:
\begin{equation}
\mathbf{Q}_{k_{i,j}}^{(h,w,m,n)} := s_{\cos}(\mathbf{H}_{k_i}^{(h,w)}, \mathbf{H}_{k_j}^{(m,n)}).
\end{equation}

The projection head is trained to align $ \mathbf{Q}_{k_{i,j}} $ with $ \mathbf{A}_{k_{i,j}} $ using the following loss:
\begin{equation}\label{eq:corr_loss}
    \mathcal{L}_{\text{corr}}(\mathbf{X}_{k_i}, \mathbf{X}_{k_j}) 
    = - \sum_{h,w,m,n} \left( \mathbf{A}_{k_{i,j}}^{(h,w,m,n)} - b \right) \mathbf{Q}_{k_{i,j}}^{(h,w,m,n)},
\end{equation}
where $ b \in \mathbb{R} $ suppresses weak correspondences. This objective encourages the segmentation head to enhance semantic coherence already present in the VFM features. After each parameter update minimizing Eq.~\ref{eq:corr_loss}, the extracted features are utilized in Eq.~\ref{eq:cluster_loss} to update the local centroid matrix $\mathbf{M}_k$, with the final local loss:
\begin{equation}
    \mathcal{L}_{\text{local}} = \mathcal{L}_{\text{corr}} + \mathcal{L}_{\text{cluster}}.
\end{equation}

This architectural design offers crucial advantages for Federated Learning. First, by restricting optimization to a lightweight segmentation head, we drastically reduce the computational burden on clients and minimize the communication overhead, as only the small projection head $S_{k}$ and centroid matrix $\mathbf{M}_k$ need to be transmitted. Second, this formulation is mathematically ideal for the federated setting because it explicitly produces discrete cluster centroids. These centroids act as semantically rich prototypes that geometrically summarize the local data distribution of each client. In our proposed methodology, these prototypes serve as a ``common language'' or set of anchor points, enabling the server to align discovered semantic classes across heterogeneous clients in a robust and efficient manner.

\begin{algorithm}[!t]
\DontPrintSemicolon
\SetKwInOut{Input}{Input}
\SetKwInOut{Output}{Output}

\Input{Clients $\mathcal{K}$, pretrained VFM $E_{\text{VFM}}$, projection head $S_k$, centroids $\mathbf{M}_k$, aggregation rounds $R$, correlation learning rate $\eta_{corr}$, centroid learning rate $\eta_{clust}$}
\Output{Global segmentation head $\bar{\theta}_S$ and centroids$\bar{\mathbf{M}}$}

\For{$r = 1$ \KwTo $R$}{
    \ForEach{client $k \in \mathcal{K}$ \textbf{in parallel}}{
        Receive $(\bar{\theta}_S, \bar{\mathbf{M}})$ from server\;
        Initialize $\theta_{S_k} \leftarrow \bar{\theta}_S$, $\mathbf{M}_k \leftarrow \bar{\mathbf{M}}$\;
        \For{each local step $i$}{
            Sample mini-batch $\{\mathbf{X}_{k_i}\}$\;
            Extract features: $\mathbf{Z}_{k_i} = E_{\text{VFM}}(\mathbf{X}_{k_i})$\;
            $\mathbf{H}_{k_i} = S_k(\mathbf{Z}_{k_i}, \theta_{S_k})$\;
            
            \tcp{Step 1: Compute Correlation loss}
            Compute $\mathcal{L}_{\text{corr}}$ using Eq. (10)\;
            
            \tcp{Step 2: Compute Cluster Loss}
            Compute $\mathcal{L}_{\text{cluster}}$ using detached features $\mathbf{H}_{k_i}$\; in Eq. (5)\;

            \tcp{Step 3: Update Parameters}
            $\theta_{S_k} \leftarrow \theta_{S_k} - \eta_{corr} \nabla_{\theta_{S_k}} \mathcal{L}_{\text{corr}}$\;
            $\mathbf{M}_k  \leftarrow \mathbf{M}_k - \eta_{clust}\nabla_{\mathbf{M}_k}\mathcal{L}_{\text{cluster}}$
        }
        Send $(\theta_{S_k}, \mathbf{M}_k)$ to server\;
    }
    Server aggregates $(\theta_{S_k}, \mathbf{M}_k) , \text{with } k=1,...,|\mathcal{K}|,$ into $(\bar{\theta}_S, \bar{\mathbf{M}})$\;
    Server broadcasts $(\bar{\theta}_S, \bar{\mathbf{M}})$ to all clients\;
}
\caption{Federated Unsupervised Semantic Segmentation (FUSS)}
\label{algo:FUSS}
\end{algorithm}

% \section{FedCC}
\subsection{Federated Aggregation Strategies}
\label{subsec:aggregation}
A major contribution of this work is the design of novel aggregation strategies for federated unsupervised semantic segmentation. In FUSS, each client transmits a local segmentation head $\theta_{S_k}$ and a centroid matrix $\mathbf{M}_k$, representing learned class prototypes. Existing FL methods are ill-suited for this setting, as they were not designed to aggregate both feature extractors and latent semantic clusters in the absence of labels.

To address this gap, we propose a suite of aggregation mechanisms tailored to the dual objective of FUSS: maintaining feature consistency across clients while achieving global alignment of semantic prototypes. These include principled adaptations of classical methods as well as entirely new strategies, specifically formulated to support end-to-end, label-free training under both i.i.d. and non-i.i.d. client distributions. To our knowledge, this is the first work to explicitly formulate aggregation objectives that jointly optimize pixel-level representation alignment and global semantic clustering in a fully federated, unsupervised semantic segmentation setting.

\subsubsection{Extended Federated Averaging}
First we extend the canonical Federated Averaging algorithm \cite{mcmahan2017communication} to our setting by aggregating both the segmentation head parameters and the centroid matrices. While standard FedAvg aggregates encoder weights, it does not account for latent class prototypes, a critical element in FUSS. Let  $\theta_{S_k}$ denote the segmentation head parameters and $\mathbf{M}_k \in \mathbb{R}^{|\mathcal{C}| \times D}$ the centroid matrix from client $k$. The aggregated global parameters are:
\begin{align}
    \bar{\theta}_S &= \sum_{k \in \mathcal{K}} \alpha_k \theta_{S_k}, \label{eq:fedavg_encoder} \\
    \bar{\mathbf{M}} &= \sum_{k \in \mathcal{K}} \alpha_k \mathbf{M}_k, \label{eq:fedavg_centroids}
\end{align}

where $  \alpha_k = \frac{N_k}{N}  $ reflects client weighting based on local dataset size, or $  \alpha_k = \frac{1}{|\mathcal{K}|}  $ for uniform aggregation. This extension serves as a simple yet effective baseline for jointly aligning learned features and semantic prototypes across clients.

\subsubsection{Centroid Clustering for Prototype Alignment (FedCC)}
While FedAvg serves as a simple baseline, it implicitly assumes that each centroid index $c$ corresponds to the same semantic class across clients. This assumption fails in unsupervised settings, where clients observe unknown subsets of the global class distribution and learn centroids in arbitrary orders. As a result, averaging centroids $\mathbf{m}_c^{i}$ and $\mathbf{m}_c^{j}$ from different clients may collapse distinct semantics into uninformative or ambiguous prototypes. This motivates a need for aggregation strategies that reason explicitly over the structure of the centroid space rather than relying on index alignment.

We propose \textbf{Fed}erated \textbf{C}entroid \textbf{C}lustering (\textbf{FedCC}), a family of novel aggregation strategies that cluster the union of local prototypes across clients. Rather than assuming consistent object-to-centroid mappings across clients, FedCC clusters the pooled centroids globally by treating all received centroids as unordered semantic candidates. This strategy remains agnostic to local class semantics and promotes emergent alignment through global structure discovery in prototype space.
% without relying on consistent class-to-centroid mappings.

 To construct the new global prototype matrix $  \bar{\mathbf{M}}  $, we pool all centroids across clients:
\begin{equation}
    \mathcal{M}_{\text{pool}} = \bigcup_{k \in \mathcal{K}} \{ \mathbf{m}_{k_c} \mid c = 1, \dots, |\mathcal{C}|\} \in \mathbb{R}^{(|\mathcal{K}| \cdot |\mathcal{C}|) \times D}.
\end{equation}

We now describe two strategies for deriving $\bar{\mathbf{M}}$ from this pool, where each row $\bar{\mathbf{m}}_c$ represents the final global semantic centroid for class $c$.

% \vspace{-2pt}
\paragraph{FedCC - $k$Means Clustering}
In this variant, the server applies $  k  $-means clustering to $  \mathcal{M}_{\text{pool}}  $, using $  C = |\mathcal{C}|  $ clusters to match the number of semantic prototypes. Each cluster center is used to construct a row in the updated global prototype matrix:
\begin{equation}
    \bar{\mathbf{m}}_c = \frac{1}{|\mathcal{C}_c|} \sum_{\mathbf{m} \in \mathcal{C}_c} \mathbf{m}, \quad \forall c = 1, \dots, |\mathcal{C}|,
\end{equation}
where $  \mathcal{C}_c \subset \mathcal{M}_{\text{pool}}  $ denotes the set of centroids assigned to cluster $  c  $. This method promotes alignment by grouping structurally similar client prototypes in the embedding space, even when local class semantics are mismatched.

% \vspace{-2pt}
\paragraph{ FedCC - Maximin Prototype Selection}
To encourage semantic diversity, we propose a maximin selection strategy. Starting with a randomly chosen centroid, each subsequent prototype is selected to maximize its minimum distance to all previously selected ones:
\begin{equation}\label{eq:maxmin_dist}
    \bar{\mathbf{m}}_c = \arg\max_{\mathbf{m} \in \mathcal{M}_{\text{pool}}} \min_{c' < c} \left\| \mathbf{m} - \bar{\mathbf{m}}_{c'} \right\|_2.
\end{equation}
This strategy avoids prototype collapse and enforces maximal separation in the feature space, which is particularly beneficial when classes appear as small objects or are dominated by background regions---conditions that often yield overlapping pixel-level representations.

In both variants, $  \bar{\theta}_S  $ is updated via Eq.~\ref{eq:fedavg_encoder}, and the resulting pair $  (\bar{\theta}_S, \bar{\mathbf{M}})  $ is broadcast to all clients. The two variants of FedCC serve complementary goals: FedCC–$k$Means promotes semantic consensus by averaging structurally similar centroids across clients, while FedCC–Maximin promotes semantic diversity, selecting centroids that span the prototype space. The former is suitable when clients share overlapping classes with misaligned feature clusters, whereas the latter guards against prototype collapse 
% in highly heterogeneous or   # COULD I DO AN EXTREME HETEROGENEOUS? EXPEriment?
class-imbalanced settings.

Figure~\ref{fig:fuss_overview} provides a visual overview of the proposed \textbf{FUSS} framework. 

\subsection{Theoretical Guarantees and Convergence Behavior}

 While FedAvg remains a cornerstone of many Federated Learning algorithms, even in modified forms, we question whether it can be made more accurate and reliable in situations where it underperforms. The FUSS setting poses a unique technical challenge: in the absence of labels or supervision, clients must align high-dimensional, semantically meaningful features across heterogeneous and potentially disjoint data distributions. Without supervision to guide inter-client alignment, naive averaging of learned class prototypes may fail, especially under non-i.i.d. data, by merging distinct classes or collapsing ambiguous regions.

While proving convergence in the fully unsupervised and federated setting remains a complex open problem currently being investigated by the research community, we offer the following theoretical justification based on known results for FedAvg and our prototype aggregation design.

Let 
\begin{equation}
    \mathbf{M}_i = \big[\mathbf{m}_i^{(1)}, \dots, \mathbf{m}_i^{|\mathcal{C}|}\big] \in \mathbb{R}^{|\mathcal{C}| \times d}
\end{equation}
denote the class prototype matrix (centroids) computed by client $i$, where $|\mathcal{C}|$ is the number of semantic clusters and $d$ the feature dimension. FedAvg-style aggregation computes the global prototype per class via:
\begin{equation}
\bar{\mathbf{m}}^{(c)} = \frac{1}{N} \sum_{i=1}^N \mathbf{m}_i^{(c)},
\end{equation}
which corresponds to the class $c$ average across $N$ clients.

\paragraph{Aligned Case}
We first argue that in cases of correct semantic alignment and global prototype convergence through FedAvg, replacing centroid averaging with FedCC does not introduce instability and will also converge. Suppose that all clients produce consistent class prototypes, i.e.,
\begin{equation}
\mathbf{m}_1^{(c)} \approx \mathbf{m}_2^{(c)} \approx \dots \approx \mathbf{m}_N^{(c)}.
\end{equation}

In this case, FedAvg and FedCC (which, in the case of Maximin, selects a maximally diverse prototype among clients) both yield similar and semantically consistent centroids:
\begin{equation}
\mathbf{m}^{(c)}_{\text{FedCC}} \approx \bar{\mathbf{m}}^{(c)}.
\end{equation}

\paragraph{Misaligned Case}
If clients produce misaligned or collapsed prototypes, averaging may produce ambiguous or meaningless centroids:
\begin{equation}
\max_{i,j} \left\| \mathbf{m}_i^{(c)} - \mathbf{m}_j^{(c)} \right\| \gg 0 
\end{equation}
that would lead $\bar{\mathbf{m}}^{(c)}$ to lie in-between distinct classes / be a collapsed representation.
In contrast, FedCC-Maximin selects the prototype $\mathbf{m}^{(c)}_{\text{FedCC}}$ that maximizes its minimum distance to all previously selected ones as shown in Eq.~\ref{eq:maxmin_dist}, pushing towards greater inter-class diversity.

On the other hand, while FedCC-kMeans will still be problematic due to the averaging nature of the algorithm, will at least find structurally consistent clusters to help avoid collapsed aggregated prototype representations.

\paragraph{Implication for Convergence}
FedCC does not interfere with the DNN weight averaging step of FedAvg (used identically in FUSS), whose convergence under standard assumptions is supported by prior work mainly through empirical results~\cite{li2020federated}. Given that the centroid aggregation step in FedCC either:
\begin{itemize}
    \item matches FedAvg when prototypes are aligned \emph{thus inherits convergence},
    \item or \emph{improves semantic consistency when FedAvg fails} (without altering weight optimization),
\end{itemize}
we argue that \emph{FedCC introduces no instability}. In fact, it enhances robustness against semantic collapse \emph{without modifying the underlying convergence properties of the global model}. 

The key technical role of FedCC can be summarized as follows: In cases where FedAvg performs well, FedCC matches its convergence behavior. However, in cases where FedAvg produces suboptimal class prototypes via naive centroid averaging, FedCC consistently improves performance by guiding local training toward more discriminative and stable semantic clusters.

\section{Experiments}\label{sec:experiments}
 
We evaluate FUSS under realistic federated scenarios involving decentralized, unlabeled datasets, without access to class labels or distributional metadata. Our goal is to measure segmentation performance under privacy-preserving constraints.
Algorithm~\ref{algo:FUSS} outlines the full FUSS training framework.

\subsection{Experimental Setup}
 
The proposed FUSS framework is evaluated under a client-server federated learning paradigm, where each client independently trains on private, non-disclosable image datasets and periodically participates in server-coordinated model aggregation rounds. Training is conducted for simulated federations with different numbers of clients, with 10 global aggregation rounds. 
In each round, all clients perform local training over their datasets and return updated parameters and centroids for aggregation. Global validation is performed centrally on a held-out set.

We conduct experiments on the following benchmarks: Cityscapes \cite{Cordts2016Cityscapes}, a 1/21 subsampled version of CocoStuff \cite{caesar2018coco}, and a private industrial dataset referred to as IPS (Industrial Pipeline Segmentation) \cite{psarras2024unified}. 

\subsubsection{Cityscapes} We employ the Cityscapes dataset to simulate cross-silo federation: we reserve all samples from three cities for validation, and partition the remaining cities among clients. Depending on the federation size (3, 6, or 18 clients), each client is assigned images from 6, 3, or 1 city respectively, promoting spatial and semantic locality in client data.

 Each federation client $k$ receives a disjoint subset $\mathcal{D}_k$ composed of images from geographically distinct cities, introducing domain distribution shifts. An example of the Cityscapes split for each federation simulation can be observed in Table \ref{tab:scenarios}. We also apply a five-crop on each image as a data augmentation technique, to better handle training times and small objects. It is important to note that while class distributions remain approximately i.i.d. across clients, the data still emulate a realistic federated learning setup in which each silo originates from a unique geographic location, thereby inducing domain and background variability among clients.
 
\begin{table}[!h]
    \renewcommand{\arraystretch}{1.3}
    \caption{Cityscapes benchmark split for applied cross-silo federation scenarios}
    \centering
    \begin{tabular}{ p{3.0cm} |  p{2.0cm} | p{2.0cm} |  p{2.0cm}}

    \toprule
    \textbf{City} &  \multicolumn{3}{c}{Configurations} \\
    \cline{2-4}
     &   3 clients & 6 clients & 18 clients\\
    \hline
    \hline
    aachen & \multirow{6}{*}{client 1} & \multirow{3}{*}{client 1} & client 1 \\
    \cline{1-1} \cline{4-4}
    bochum &  &  & client 2 \\
    \cline{1-1} \cline{4-4}
    bremen &  &  & client 3 \\
    \cline{1-1} \cline{3-4}
    cologne &  & \multirow{3}{*}{client 2} & client 4 \\
    \cline{1-1} \cline{4-4}
    darmstadt &  &  & client 5 \\
    \cline{1-1} \cline{4-4}
    dusseldorf &  &  & client 6 \\
    \cline{1-1} \cline{2-2} \cline{3-4}
    erfurt & \multirow{6}{*}{client 2} & \multirow{3}{*}{client 3} & client 7 \\
    \cline{1-1} \cline{4-4}
    hamburg &  &  & client 8 \\
    \cline{1-1} \cline{4-4}
    hanover &  &  & client 9 \\
    \cline{1-1} \cline{3-4}
    jena &  & \multirow{3}{*}{client 4} & client 10 \\
    \cline{1-1} \cline{4-4}
    krefeld &  &  & client 11 \\
    \cline{1-1} \cline{4-4}
    monchengladbach &  &  & client 12 \\
    \cline{1-1} \cline{2-2} \cline{3-4}
    stuttgart & \multirow{6}{*}{client 3} & \multirow{3}{*}{client 5} & client 13 \\
    \cline{1-1} \cline{4-4}
    strasbourg &  &  & client 14 \\
    \cline{1-1} \cline{4-4}
    tubingen &  &  & client 15 \\
    \cline{1-1} \cline{3-4}
    ulm &  & \multirow{3}{*}{client 6} & client 16 \\
    \cline{1-1} \cline{4-4}
    weimar &  &  & client 17 \\
    \cline{1-1} \cline{4-4}
    zurich &  &  & client 18 \\
    \bottomrule
    \end{tabular}
    \label{tab:scenarios}
\end{table}

\subsubsection{CocoStuff} Although the CocoStuff dataset is not inherently designed for real-world federated learning scenarios, it is employed here to evaluate the robustness of our framework under extreme inter-client class heterogeneity. Each image is assigned a dominant semantic label based on the pixel-wise frequency of classes, after which a Dirichlet-based sampling strategy is applied to partition the dataset among clients. This procedure induces a controlled yet pronounced class imbalance and variability across clients.

We construct non-i.i.d. partitions of the CocoStuff dataset using a Dirichlet-based sampling strategy informed by semantic supercategories. The process involves assigning each image to a single coarse semantic class based on pixel frequency, followed by Dirichlet sampling to allocate images to clients in a skewed but controllable manner.

\paragraph{Dominant Supercategory Assignment}
Each CocoStuff image is annotated with a pixel-wise segmentation mask covering 182 fine-grained classes. To facilitate class-conditional sampling, we assign a single coarse label to each image based on the most frequently occurring supercategory. Let $  \mathbf{Y}' \in \mathbb{R}^{H \times W}  $ be the RGB fine-grained segmentation mask of an image $\mathbf{X}$, and let $  f : \mathcal{C}_\text{fine} \to \mathcal{C}_\text{coarse}  $ be a known mapping from the 182 fine-grained to the 27 coarse classes that we use in our experiments, following common USS conventions. The dominant coarse label for image $  \mathbf{X}  $ is defined as:
\begin{equation}
  \hat{c} = \arg\max_{c \in \mathcal{C}_\text{coarse}} \left| \left\{ (h, w) \mid f\left( \mathbf{Y}'^{(h,w)} \right) = c \right\} \right|,  
\end{equation}
i.e., the supercategory $  \hat{c}  $ that occurs most frequently across all pixels of the image. Each image is thus labeled with a single dominant supercategory, enabling label-driven data partitioning despite the absence of ground-truth image-level annotations.
 
Once each image is tagged with its dominant coarse category, we simulate heterogeneous federated settings by partitioning the data using Dirichlet sampling over coarse class distributions. Specifically, for $  K  $ clients and a Dirichlet concentration parameter $  \alpha  $, we draw a probability vector $  \mathbf{p}_c \sim \text{Dir}(\alpha \cdot \mathbf{1}_K)  $ for each supercategory $  c \in \mathcal{C}_\text{coarse}  $. This vector determines the probability that an image with dominant category $  c  $ is assigned to each client.

\begin{figure}[!htbp]
  \centering
  % First subfigure
  \begin{subfigure}[b]{\linewidth}
    \centering
    \includegraphics[width=1.1\linewidth]{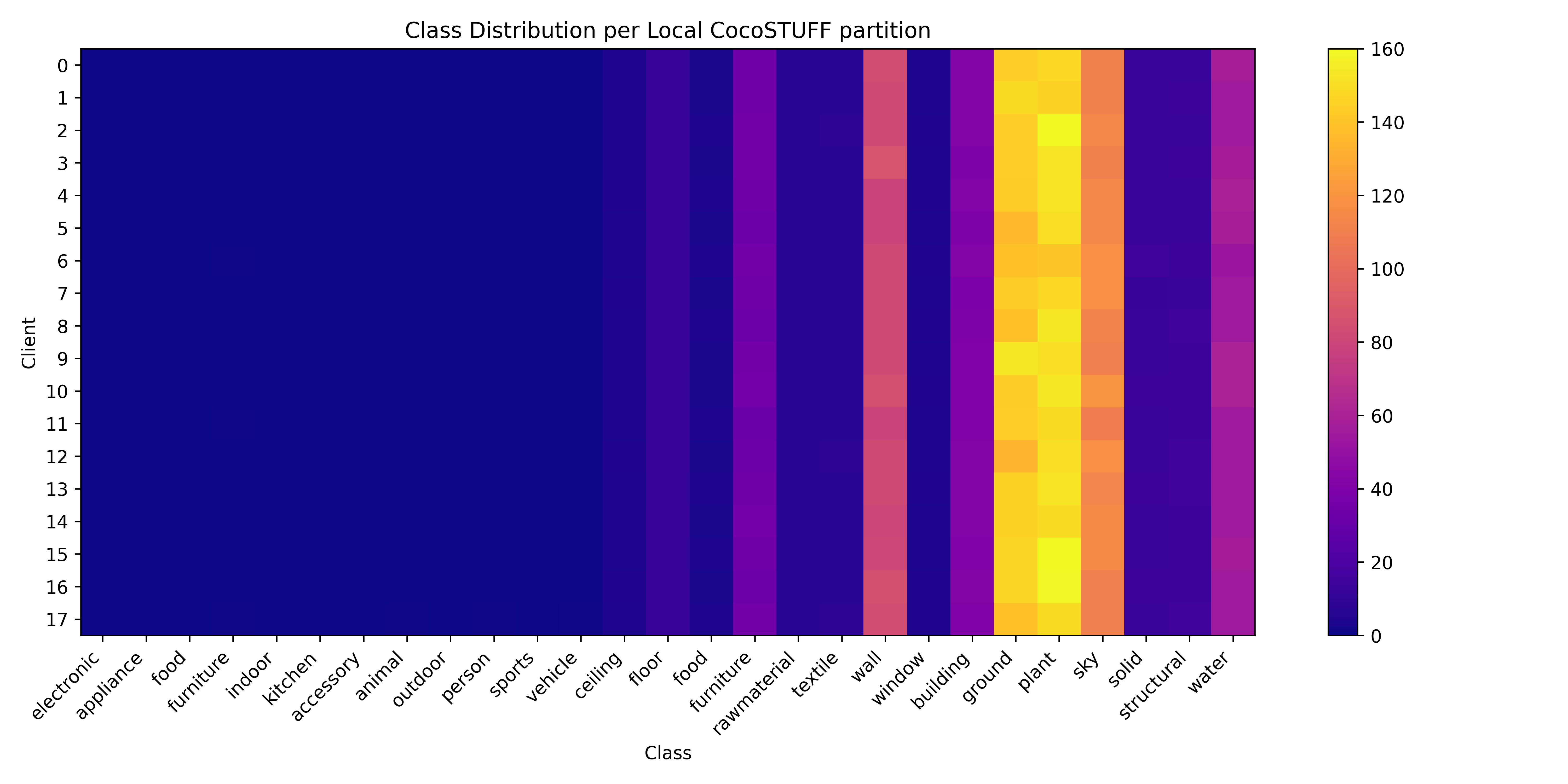}
    \caption{ }
    \label{fig:iid}
  \end{subfigure}
  % \vspace{1em}  % Adds vertical spacing

  % Second subfigure
  \begin{subfigure}[b]{\linewidth}
    \centering
    \includegraphics[width=1.1\linewidth]{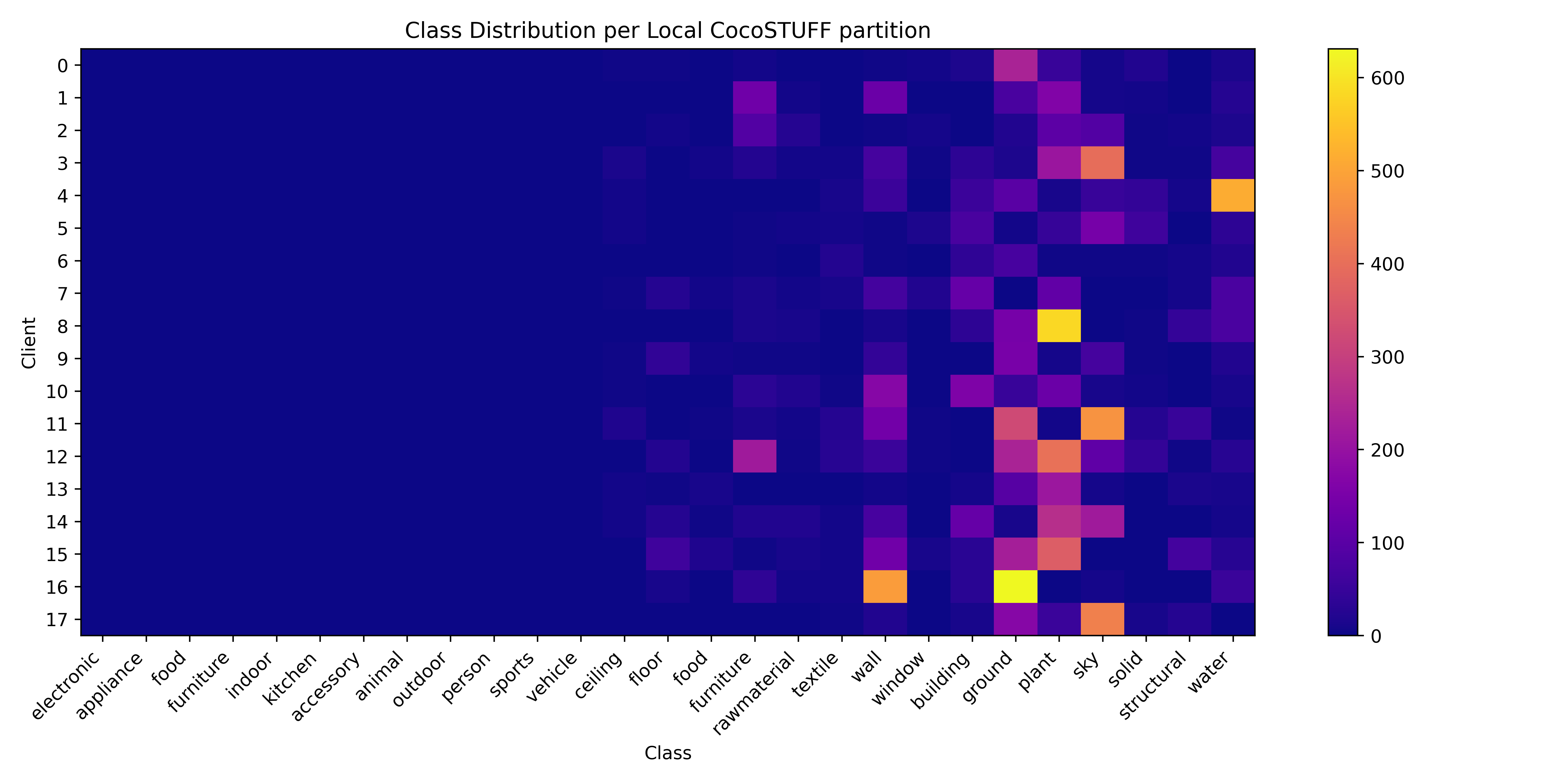}
    \caption{}
    \label{fig:non-iid}
  \end{subfigure}

  \caption{CocoStuff split visualization for 18 clients, based on dominant class frequency. (a) i.i.d. partition where dominant classes are distributed uniformly. (b) non-i.i.d. partition with Dirichlet concentration parameter $\alpha = 0.5$, where samples are concentrated on specific clients. Note the varying frequency scales (right) indicating the degree of sample concentration at each client.}
  \label{fig:proto_alignment}
\end{figure}

Images are then distributed across clients accordingly. Lower values of $  \alpha  $ (e.g., 0.5) induce higher heterogeneity by skewing the distribution of dominant classes across clients, while larger $  \alpha  $ values produce more balanced partitions. In Figure~\ref{fig:proto_alignment}, we provide a visual comparison of prototype alignment under both i.i.d. and non-i.i.d. client distributions.

This approach ensures that each client receives a diverse but biased subset of the full dataset, approximating realistic non-i.i.d. scenarios in federated learning while retaining control over the degree of distributional shift.

\subsubsection{IPS} To evaluate real-world applicability, we have employed the IPS \cite{psarras2024unified} dataset, which contains high-resolution RGB images and video frames captured at four industrial sites performing pipeline infrastructure inspection. IPS poses a binary semantic segmentation task with two classes: \texttt{background} and \texttt{pipe}, and contains 984 training samples. We simulate a 3-client federation by assigning each client to a distinct site, while a fourth, unseen site is reserved exclusively for validation---assessing generalization under cross-domain deployment.

\subsection{Implementation Details}\label{subsec:implementation}
 
Each client is equipped with a frozen Vision Foundation Model (VFM) backbone, specifically the base version of DINO \cite{caron2021emerging} pretrained on ImageNet \cite{russakovsky2015imagenet}, which is used to extract semantically rich features from raw input images. On top of this backbone, a trainable segmentation head $S(\cdot, \theta_S)$ is implemented as a lightweight two-layer convolutional projection network that maps the VFM 768-dimensional pixel embeddings, to a lower-dimensional
% , clustering-compatible 
embedding space of 70 dimensions.

Training of each local segmentation head $S_k$ is performed using mini-batches of 8 query images sampled from the client's dataset. For each query image, we select one nearest neighbor (by feature similarity) and five additional random support images from the same client. The cross-image correlation loss (Eq.~\ref{eq:corr_loss}) is computed between the query and each support image individually. Loss hyperparameters $\lambda$ and $b$ are typically discovered manually, via grid-searching from a predefined set of values in the range $[0,1]$ and cross-validating on an independent annotated test set. In binary segmentation cases \cite{tzimas2026ews}, apart from heuristic methods, optimal values can be automatically approximated through Gaussian Mixture Models. In our experiments, we utilize the established hyperparameters from \citep{hamilton2022unsupervised, kim2024eagle, hahn2024boosting} to ensure a fair baseline configuration and ensure fairness in comparisons with prior works. We use the Adam optimizer \cite{kingma2014adam} with a learning rate of $5 \times 10^{-4}$ to optimize $\theta_{S_k}$. prior works.

Following each local update, centroid optimization is performed by clustering the combined batch of query and support feature vectors into $|\mathcal{C}| = 27$ clusters for Cityscapes and CocoStuff, or $|\mathcal{C}| = 2$ for IPS. After computing the clustering loss from Eq.~\ref{eq:cluster_loss}, centroids are updated using a second Adam optimizer with a learning rate of $5 \times 10^{-3}$.

All clients perform an equal number of local updates, configured to ensure that each client completes at least one full epoch over its local dataset. While the number of local steps remains fixed within each experiment, it may vary across experiments depending on available computational resources and batch size configurations. Upon completing local training, clients transmit their updated parameters to the server, which then applies one of the aggregation strategies described in Section~\ref{subsec:aggregation} and broadcasts the resulting global model back to all clients.

\subsection{Results}

We evaluate FUSS on both standard benchmarks (Cityscapes) and a real-world deployment scenario (IPS), under a range of federation configurations. The primary evaluation metric is Mean Intersection-over-Union (mIoU), computed on held-out validation sets. Ground-truth labels are never used during training and serve strictly for evaluation.

To provide broader context, we also incorporate two common local training regularizers---FedProx \cite{li2020federated} and FedMoon \cite{li2021model}---which are orthogonal to our aggregation strategies but useful for understanding FedCC’s modularity and robustness under both i.i.d. and non-i.i.d. conditions.

\paragraph{FedProx} FedProx \cite{li2020federated} adds a proximal term to the local loss to prevent divergence from the global model under non-i.i.d. data. In FUSS, we apply it to the segmentation head:
\begin{equation}\label{eq:fedprox}
\mathcal{L}_{\text{FedProx}} = \mathcal{L}_{\text{local}} + \frac{\mu}{2} \left\| \theta_{S_k} - \bar{\theta}_S \right\|^2,
\end{equation}
where $  \mu  $ controls regularization strength. Since encoders are frozen and centroids are directly aggregated, only the segmentation head is regularized.

\paragraph{FedMoon} FedMoon \cite{li2021model} introduces a contrastive loss to align local and global feature representations. In FUSS, we apply this to the segmentation embeddings:
\begin{equation}
   \mathcal{L}_{\text{Moon}} = - \log \left( \frac{ \exp( s_{\cos}(\mathbf{z}_k, \bar{\mathbf{z}})/\tau ) }{ \exp( s_{\cos}(\mathbf{z}_k, \bar{\mathbf{z}})/\tau ) + \exp( s_{\cos}(\mathbf{z}_k, \mathbf{z}^{\text{prev}}_k)/\tau ) } \right), 
\end{equation}
and define the final objective as $  \mathcal{L}_{\text{FedMoon}} = \mathcal{L}_{\text{local}} + \lambda \cdot \mathcal{L}_{\text{Moon}}  $ with $\lambda$ a tunable hyperparameter. This promotes representation consistency across rounds.

\begin{table*}[!htbp]
    \renewcommand{\arraystretch}{1.2}
    \caption{Experimental results comparing naive and FedCC versions of FedAvg, FedProx, and FedMoon across different client counts. Reported metric is mean Intersection-over-Union (mIoU). CRF post processing is applied to all results.}\label{tab:miou_iid}
    \centering
    \begin{tabular}{l | c | c |c| c }
        \toprule
        \textbf{Method}& \multicolumn{3}{c|}{\textbf{Cityscapes}} &{ \textbf{IPS}} \\
        \cline{2-5}
         & 3 Clients & 6 Clients & 18 Clients & 3 Clients  \\
         
        \hline
        \hline
        FedProx
             & 19.77  & 20.01  & 18.22 & \textbf{84.63}  \\
        \textbf{FedProx+FedCC}
             & \textbf{20.96} & \textbf{20.48} & \textbf{20.90}  & 83.52 \\
        \hline
        FedMOON
             & 19.83 & 19.94 & 19.86  &  83.78 \\
        \textbf{FedMOON+FedCC}
             & \textbf{20.35} & \textbf{21.76} & \textbf{20.41}  & \textbf{84.70} \\
        \hline
        FedAvg
             & 21.12  & 19.94 & 21.25 &  84.10 \\
        \textbf{FedAvg+FedCC}
             & \textbf{21.83} & \textbf{21.80} & \textbf{21.53} & \textbf{84.87} \\
        \hline
        \hline
        Centralized & \multicolumn{3}{c|}{21.0}  &   {85.66}   \\
        \bottomrule
    \end{tabular}
\end{table*}

Table~\ref{tab:miou_iid} reports segmentation results for FedAvg, FedProx, and FedMoon, both with and without FedCC. Since settings remain constant across all methods and client counts, FedCC remains the only differentiating factor for the improved segmentation performance. Notably, FedAvg+FedCC achieves 21.72 mIoU on average on all federated scenarios for Cityscapes and 84.87 mIoU on IPS, closely matching centralized upper bounds. 

A subtle performance discrepancy is observed in the FedProx + FedCC configuration on the IPS dataset compared to the baseline. This is attributed to the highly imbalanced sample distribution of the IPS dataset, where a single majority client
holds $\approx80\%$ of the training images. In this setting, under weighted parameter aggregation, the proximal term in FedProx (Eq. \ref{eq:fedprox}) creates a negative bias loop: at each aggregation round, smaller clients are increasingly penalized for diverging from the majority-biased global model, suppressing site-specific features and reducing required diversity. Hence, for baseline FedProx on IPS, we report the the unweighted aggregation configuration, to avoid reporting results skewed by this negative bias loop. In contrast, we maintain a fixed protocol for all variations of FUSS when augmented with FedCC, using weighted aggregation across all datasets, for empirical rigor and to demonstrate generalizability \textbf{without the need for site-specific tuning}.

\begin{table}[!htbp]
% \begin{table}
    \caption{Comparison of FedCC performance, with FUSS applied to non-i.i.d (Dirichlet, $\alpha=0.5$) CocoStuff benchmark, for 18 clients.}
    \label{tab:iid}
   \centering
    \renewcommand{\arraystretch}{1.2}
    \begin{tabular}{l|p{1.2cm}}
    \toprule
    \textbf{Method} & \textbf{mIoU} \\
        \hline
        \hline
        % \textbf{Category} & \textbf{IPS (3 Clients)} \\
        FedProx & 22.39 \\
        FedMoon & 24.69 \\
        FedAvg & 24.69 \\
        \hline
        FedProx + FedCC & 24.38 \\
        FedMoon + FedCC & 23.15 \\
        \textbf{FedAvg + FedCC} & \textbf{25.39}\\
        \hline
        \hline
        Centralized & 25.41 \\
        \bottomrule
    \end{tabular}
% \end{table}
\end{table}
To evaluate FUSS under heterogeneous client distributions, we partition CocoStuff to 18 clients, using Dirichlet sampling ($\alpha=0.5$). Table~\ref{tab:iid} shows that FedAvg+FedCC achieves 25.39 mIoU, nearly matching the centralized score of 25.41, and outperforming both standard and FedCC-enhanced versions of FedProx and FedMoon.

\begin{figure*}[!htbp]
  \centering
  \includegraphics[width=1.0\linewidth]{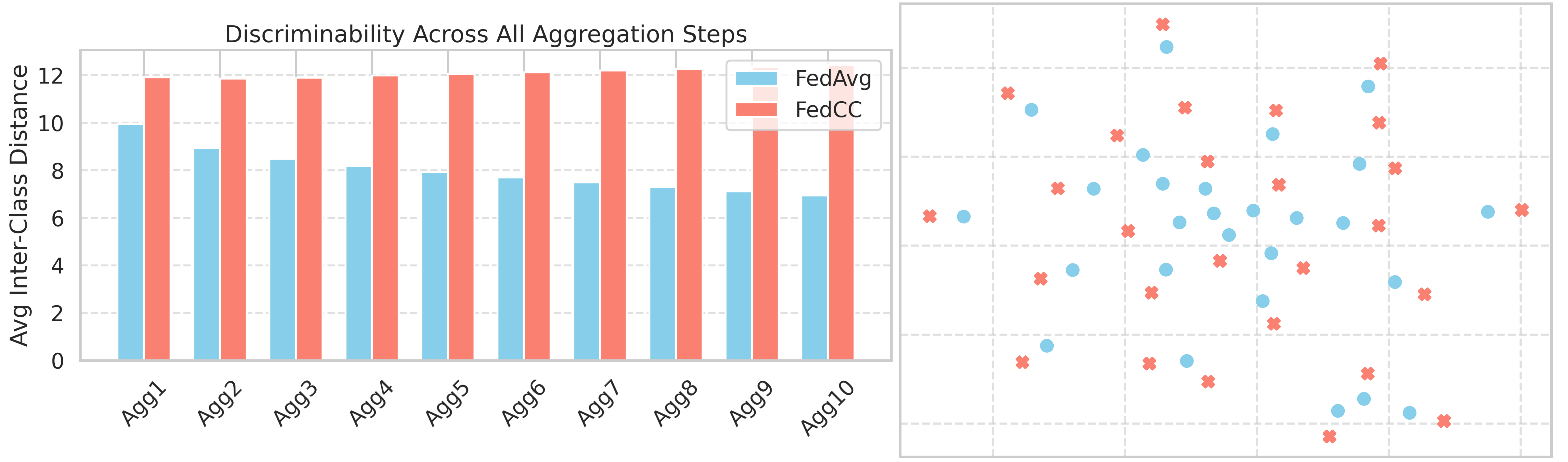}
  \caption{Left: Discriminability analysis between resulting centroids for CocoStuff non-i.i.d. Right: t-SNE projections of centroids from final federated aggregation.}
  \label{fig:qualitative}
\end{figure*}

\begin{figure*}[!htbp]
  \centering
  \includegraphics[width=1.0\linewidth]{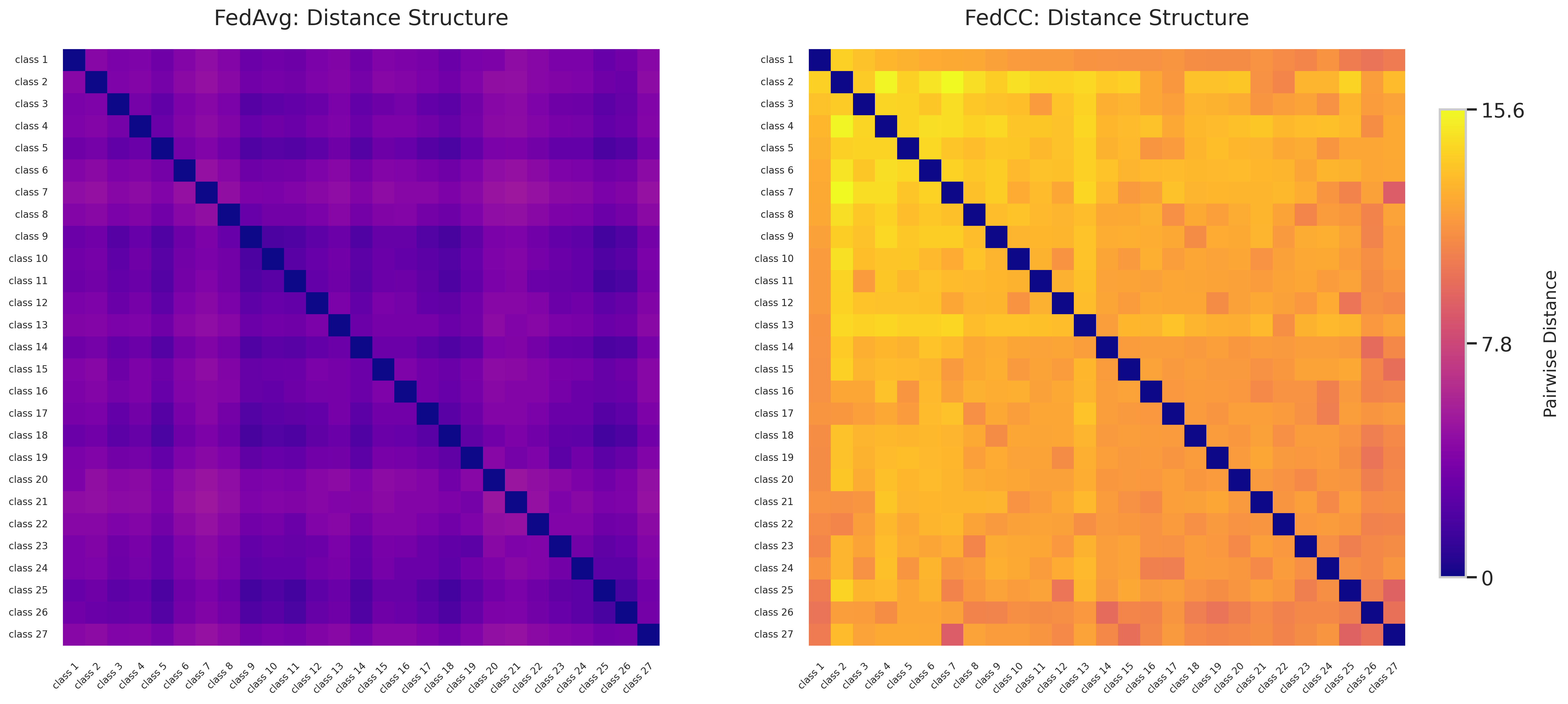}
  \caption{Pairwise distance structure for FedAvg (left) and FedAvg+FedCC (right). FedAvg+FedCC achieves greater inter-class distances across all 27 classes of CocoStuff.}
  \label{fig:pairwise_comparison}
\end{figure*}

A qualitative assessment of the results under heterogeneous data distributions is presented in Figure~\ref{fig:qualitative}. These examples illustrate that even in strongly non-i.i.d. scenarios, the centroids produced by FedCC remain structurally discriminative across the entire federation. Furthermore, visualizations of the projected centroid embedding space reveal that while FedAvg tends to produce cluttered prototypes, FedCC maintains significantly greater inter-class separability. Similar structural improvements for FedProx and FedMoon are provided in \ref{App:A}. Additionally, we visualize the pairwise inter-class distances in Figure~\ref{fig:pairwise_comparison} to demonstrate the structural separation achieved by FedCC across the entire 27-class distribution of CocoStuff, after the final aggregation round. It is important to note that while the t-SNE projections provide a qualitative 2D visualization of class grouping, the pairwise distance matrices in Figure~\ref{fig:pairwise_comparison} offer a more precise quantitative assessment. These distances are computed directly within the original 70-dimensional feature space, thereby preserving the high-dimensional structural relationships that are inevitably distorted during the non-linear dimensionality reduction required for Figure~\ref{fig:qualitative}.

To isolate the contributions of each component in FUSS, we perform an ablation study on the aggregation process, evaluating the effects of \textbf{weighted averaging} (W) of each client parameters, \textbf{encoder-only parameter fusion} (E), and \textbf{prototype-only aggregation} (C). Table~\ref{tab:ablation} presents results across multiclass  (Cityscapes) and binary (IPS) segmentation \emph{without applying CRF} \cite{krahenbuhl2011efficient} in post-processing, for computational efficiency.
% 
% \begin{table}[!bp]
\begin{table}[!htbp]
    \renewcommand{\arraystretch}{1.2}
    \caption{
       Ablation study for multiclass (Cityscapes) and binary (IPS) federated unsupervised semantic segmentation. 
    We evaluate the impact of three aggregation modules: 
    W = weighted averaging (by client dataset size), 
    E = encoder-only aggregation, 
    C = centroid-only aggregation. 
    Reported scores are mean Intersection-over-Union (mIoU). For configurations with no resulting global model, we report the average result along with the best (+) and worst(-) performing clients.
      %this makes the FedCC results even more interpretable --- because C-only aggregation becomes meaningful and outperforms *consistently* naive E-only FedAvg. That strengthens the motivation behind the prototype alignment design.
    }\label{tab:ablation}
    \centering
    \setlength{\tabcolsep}{4pt}
    \begin{tabular}{ p{1.8cm} |  p{0.8cm} | p{0.8cm} |  p{0.8cm} | p{1.7cm} | p{1.7cm} | p{1.8cm} | p{1.7cm} }
    \toprule
    \textbf{Method} & \multicolumn{3}{c|}{\textbf{Agg. Modules}} & \multicolumn{3}{c|}{\textbf{Cityscapes}} & \textbf{IPS} \\
    \cline{2-8}
     & W & E & C & 3 Clients & 6 Clients & 18 Clients & 3 Clients \\
    \hline
    \hline
    {\footnotesize Local-Only} 
     & & &     &  $18.60_{-0.69}^{+0.72}$ & $18.20^{+1.07}_{-1.62}$ & $16.74^{+1.71}_{-1.62}$ & $67.24^{+15.0}_{-27.0}$ \\
     \hline
    \multirow{6}{*}{FedAvg} 
     & &\checkmark &     &  18.55 $_{-0.7}^{+0.6}$ & $17.97^{+0.72}_{-0.48}$ & $17.18^{+1.52}_{-1.58}$ & $70.74^{+12.4}_{-18.0}$ \\
     & &          & \checkmark & $19.91_{-0.15}^{+0.24}$ & $18.11_{-0.61}^{+0.74}$ & $16.46_{-1.17}^{+0.90}$  & $80.88_{-2.4}^{+2.9}$ \\ 
     & & \checkmark & \checkmark & 19.26 & 18.22 & 17.55 & 80.64 \\
    % \hline
    % \multirow{3}{*}{Weight. FedAvg} 
     & \checkmark & \checkmark &           & 18.55 $_{-0.7}^{+0.7}$ & $18.45^{+0.47}_{-1.22}$& $17.36^{+1.42}_{-2.20}$ & $78.43^{+4.78}_{-4.73}$ \\
     & \checkmark &          & \checkmark & $18.69^{+0.60}_{-0.59}$ & $17.98^{+0.63}_{-0.85}$ & $16.51^{+0.99}_{-1.14}$ & $79.10_{-1.1}^{+1.3}$ \\ 
     \cline{2-8}
     & \checkmark & \checkmark & \checkmark & 19.24 & 18.96 & 17.50  & 79.16 \\
    \hline
    \multirow{4}{*}{\shortstack{FedCC\\$-$\\kmeans}} 
      & & &\checkmark& $18.91^{+0.67}_{-0.86}$  & $18.90^{+0.66}_{-0.83}$ & $17.35^{+1.1}_{-1.4}$ & \underline{81.14}$^{+2.76}_{-2.07}$ \\
    & \checkmark & &\checkmark& $18.61^{+0.33}_{-0.55}$  & $18.59^{+0.33}_{-0.60}$ & $17.50^{+1.0}_{-1.1}$ & $79.49^{+1.29}_{-1.11}$ \\
       & & \checkmark & \checkmark& 18.88 & \underline{19.41} & 17.89  & 80.67 \\
       \cline{2-8}
      & \checkmark & \checkmark & \checkmark & \textbf{19.71} & 19.39 & 17.87  & \textbf{82.60} \\
    \hline
    \multirow{4}{*}{\shortstack{FedCC\\$-$\\maximin}} 
      & & &\checkmark& $18.51^{+0.11}_{-0.08}$ & $18.90^{+0.66}_{-0.83}$ & $17.06^{+1.26}_{-1.45}$  & 81.01$^{+2.9}_{-1.65}$ \\
      & \checkmark & &\checkmark& $19.06^{+0.22}_{-0.45}$  & $18.57^{+0.56}_{-0.55}$  &  $17.54^{+0.9}_{-1.2}$ & 79.43$^{+1.5}_{-1.4}$ \\
      & & \checkmark & \checkmark & 19.50 & 19.38 & \underline{18.12} & 78.22 \\
      \cline{2-8}
      & \checkmark & \checkmark & \checkmark & \underline{19.52} & \textbf{19.64} & \textbf{18.87} & 80.27 \\
    \bottomrule
    % \hline
    % \multirow{2}{*}{Weight. Global + FedAvg} 
    \end{tabular}
\end{table}
The aggregation module analysis confirms the benefit of explicitly aligning prototypes (C), even without encoder averaging. For instance, C-only FedAvg outperforms E-only FedAvg across the majority of settings. On IPS, FedCC-KMeans performs best, likely due to strong foreground-background separability. However, for Cityscapes, where small objects and semantic clutter dominate, FedCC-Maximin proves more effective by promoting inter-centroid diversity and mitigating prototype collapse.

\begin{figure}[!h]
  \centering
  \includegraphics[width=0.8\linewidth]{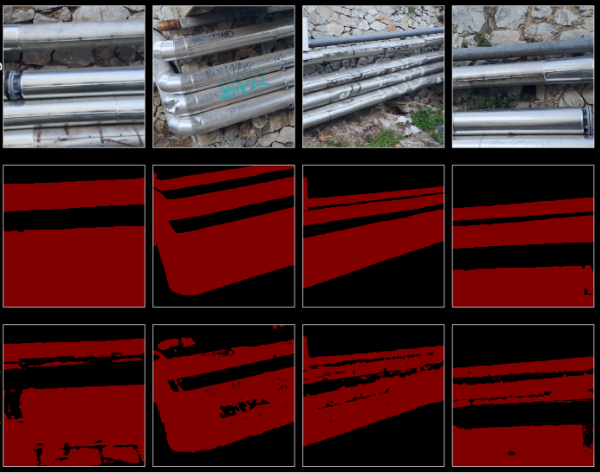}
  \caption{
  Visual overview of the IPS dataset. 
  \textbf{Top:} Sample validation images. 
  \textbf{Middle:} Corresponding ground-truth binary segmentation masks. 
  \textbf{Bottom:} Predicted masks produced by FUSS with FedCC (k-means) aggregation.
  }
  \label{fig:ips_data_results}
\end{figure}

Together, our results demonstrate that FUSS enables accurate, decentralized segmentation with minimal communication, and that our aggregation strategies scale across dataset complexities and federation settings. 

\begin{figure}[!h]
  \centering
  \includegraphics[width=0.8\linewidth]{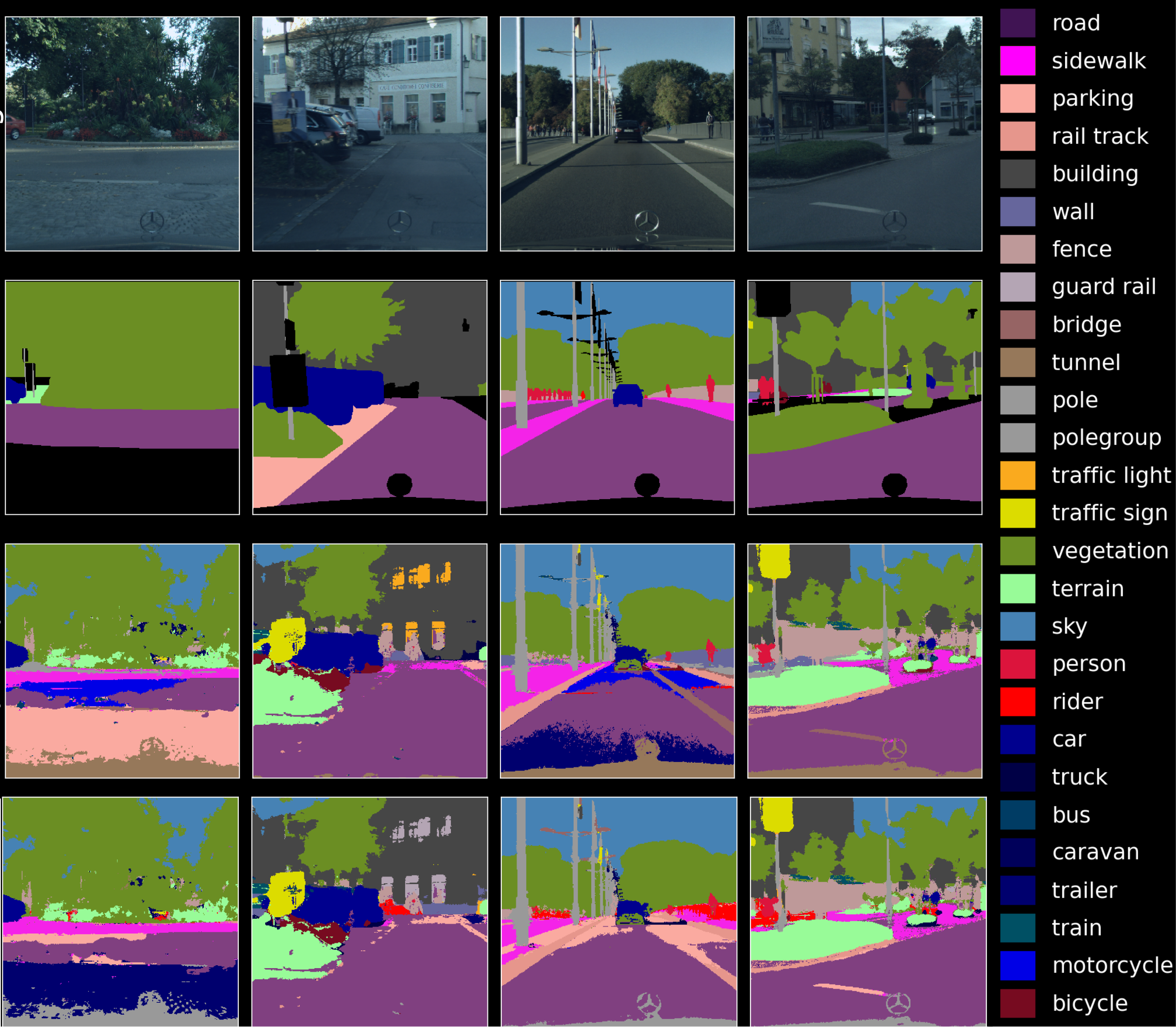}
  \caption{
  Visual representation Cityscapes results. 
  \textbf{Top:} Sample validation images. 
  \textbf{Top-Middle:} Corresponding ground-truth binary segmentation masks. 
  \textbf{Bottom-Middle:} FedAvg FUSS results.
  \textbf{Bottom:} FedCC FUSS results.
  }
  \label{fig:cityscapes_data_results}
\end{figure}

\subsection{Statistical Significance}
While our method achieves consistent improvements over strong baselines across most experimental settings, the absolute performance gains may appear modest. We consider it essential to demonstrate the consistency and statistical reliability of the proposed approach. To this end, our evaluation protocol is extended to include \emph{per-image} mIoU tracking across all test samples, enabling statistical hypothesis testing at the image level rather than relying solely on aggregated mean scores.

Specifically, we evaluated the null hypotheses: a) that the differences in the mean IoU values between the two methods are the same, using a paired t-test \cite{hsu2014paired}), and b) the median difference between two distributions is equal to zero, using the Wilcoxon signed-rank test \cite{woolson2007wilcoxon}. We conducted both tests between our proposed FedCC and FedAvg under all federation settings (3, 6, and 18 clients) and across all datasets (Cityscapes, CocoStuff, IPS).

For CocoStuff specifically, we reran the FedAvg and FedAvg+FedCC experiments with four additional seeds, as the results appeared to be the most incremental. We averaged the mIoU results across all runs and then tested these averaged values to obtain the p-value, which shows the probability the that null hypothesis of the two tests (same mean IoU values, same median IoU values). These tests were chosen to capture both mean differences and distributional shifts across matched samples (i.e., the same image evaluated under two methods). We present our results in tables \ref{tab:stats_tests_domain} and \ref{tab:stats_tests_class}.

\begin{table}[!htbp]
    \renewcommand{\arraystretch}{1.2}
    \caption{Statistical significance results for \textbf{domain non-iid} comparing FedAvg vs. FedAvg+FedCC. Reported values are $p$-values from paired t-test (parametric) and Wilcoxon signed-rank test (non-parametric). Lower $p$ indicates stronger significance.}\label{tab:stats_tests_domain}
    \centering
    % \resizebox{\textwidth}{!}{%
    \begin{tabular}{l | c | c | c | c }
        \toprule
        \textbf{Dataset / Test} & \multicolumn{3}{c|}{\textbf{Cityscapes}} & \textbf{IPS}  \\
        \cline{2-5}
         & 3 Clients & 6 Clients & 18 Clients & 3 Clients \\
        \hline
        \hline
        Paired t-test
            & $1.20 \times 10^{-04}$ 
            & $5.89 \times 10^{-28}$ 
            & $3.46 \times 10^{-05}$ 
            & $5.29 \times 10^{-31}$ \\
        \hline
        Wilcoxon signed-rank
            & $7.49 \times 10^{-07}$ 
            & $1.43 \times 10^{-25}$ 
            & $1.99 \times 10^{-03}$ 
            & $7.99 \times 10^{-29}$ \\
        \bottomrule
    \end{tabular}
    % }
\end{table}

\begin{table}[!htbp]
    \renewcommand{\arraystretch}{1.2}
    \caption{Statistical significance results for \textbf{class non-iid} comparing FedAvg vs. FedAvg+FedCC. Reported values are $p$-values from paired t-test (parametric) and Wilcoxon signed-rank test (non-parametric). Lower $p$ indicates stronger significance.}\label{tab:stats_tests_class}
    \centering
    % \resizebox{\textwidth}{!}{%
    \begin{tabular}{l | c}
        \toprule
        \textbf{Dataset / Test} & \textbf{CocoStuff (non-iid)} \\
        \cline{2-2}
         & 18 Clients \\
        \hline
        \hline
        Paired t-test
           & $1.47 \times 10^{-02}$ \\
        \hline
        Wilcoxon signed-rank
           & $4.75 \times 10^{-03}$ \\
        \bottomrule
    \end{tabular}
    % }
\end{table}

These results confirm that the improvements from FedCC are \emph{statistically significant}, and not attributable to random variation or evaluation noise. We note that the extremely low p-values (e.g., $p < 10^{-25}$) are expected due to the large number of samples and the consistent performance advantage observed across nearly all test images.  Our results provide strong empirical evidence that the performance gains of FedCC are consistent, statistically sound, and practically meaningful.

\subsection{On Practical Guidance and Applicability}
One of the contributions of our study lies in the empirical comparison of two FedCC variants---\textbf{$k$-means-based} and \textbf{Maximin-based} aggregation---across two distinct Unsupervised Segmentation tasks/datasets that vary in visual complexity and inter-client class distribution. Our findings offer concrete heuristics for when each variant is preferable:

\begin{itemize}
    \item \textbf{FedCC-$k$-means} performs better in binary or visually simple tasks such as the IPS dataset, where the camera consistently captures large, separable foreground objects (pipeline) against a more uniform background (Figure~\ref{fig:ips_data_results}). In these scenarios, the benefits of inter-class discriminability provided by Maximin are diminished, and $k$-means offers a computationally efficient and robust solution.
    
    \item \textbf{FedCC-Maximin} is more effective in complex datasets such as Cityscapes, which exhibit a wide range of object classes, diverse spatial scales, and occlusions. Here, the Maximin strategy better preserves diversity across clients and avoids prototype collapse as shown in Figure~1, leading to higher segmentation accuracy.
\end{itemize}

\section*{Limitations}\label{sec:limitations}
 
Despite its practical value, FUSS inherits key limitations from the USS paradigm. Most notably, it assumes a fixed number of semantic regions $|\mathcal{C}|$ across all clients, regardless of local content. This may lead to over-clustering in simpler clients or under-representation of fine-grained classes in complex ones. Additionally, following common FL conventions, our encoder aggregation relies on simple weighted averaging, without taking into account local model quality or encoding variance. Future work could explore adaptive strategies based on representation diversity or cluster coherence.

Finally, while we simulate non-i.i.d. settings using spatial splits (Cityscapes) and Dirichlet sampling (CocoStuff), the lack of a standardized protocol for inducing heterogeneity in unsupervised segmentation remains an open problem. Defining realistic, benchmarkable heterogeneity scenarios is an important direction to evaluate federated USS frameworks more rigorously.

\section{Conclusion}\label{sec:conclusion}
 
This work introduced \textbf{FUSS}, a Federated Unsupervised image Semantic Segmentation framework designed for decentralized, label-free semantic image segmentation under strict data privacy constraints. FUSS builds upon recent advances in self-supervised vision foundation models and clustering-based optimization to enable collaborative semantic segmentation without the need for ground-truth annotations or data sharing.

We addressed a critical gap in existing federated learning literature by proposing a general formulation that integrates unsupervised representation learning with class prototype alignment across clients. Through a combination of correlation-based local optimization and novel aggregation strategies, including variations of weighted parameter averaging and global clustering procedures, we demonstrated that FUSS effectively mitigates challenges such as semantic misalignment and local overfitting.

Extensive experiments on both benchmark (Cityscapes, CocoStuff) and real-world (IPS) datasets show that our method consistently outperforms  local-only and traditional Federated Learning training baselines, achieving segmentation quality competitive with centralized unsupervised models, while remaining scalable and robust in i.i.d. and non-i.i.d. data regimes. 
% Notably, FUSS maintains scalability across federation sizes and performs robustly 
% in low-data regimes
% in i.i.d and non-i.i.d data regimes, offering a practical and communication-efficient solution for privacy-sensitive domains such as industrial inspection and autonomous systems.
Future extensions may explore end-to-end learnable aggregation schemes, adaptive client weighting strategies based on local task uncertainty, or integration with semi-supervised constraints when limited labels become available.

\section*{Acknowledgement}
This work has received funding from the European Union Horizon Europe research and innovation programme under grant agreement number 101070604 – SIMAR. This publication reflects only the authors’ views. The European Commission is not responsible for any use that may be made of the information it contains.

\section*{Declaration of generative AI and AI-assisted technologies in the manuscript preparation process}\label{sec:declaration}
During the preparation of this work the author(s) used ChatGPT in order to do language refinement and grammatical verification. After using this tool/service, the author(s) reviewed and edited the content as needed and take(s) full responsibility for the content of the published article. 

\bibliographystyle{elsarticle-harv}
\bibliography{refs}

\clearpage
\appendix
\section{Discriminability visualizations}\label{App:A}
\begin{figure}[!h]
    \centering
    \begin{minipage}{0.48\textwidth}
        \centering
        \includegraphics[width=\linewidth]{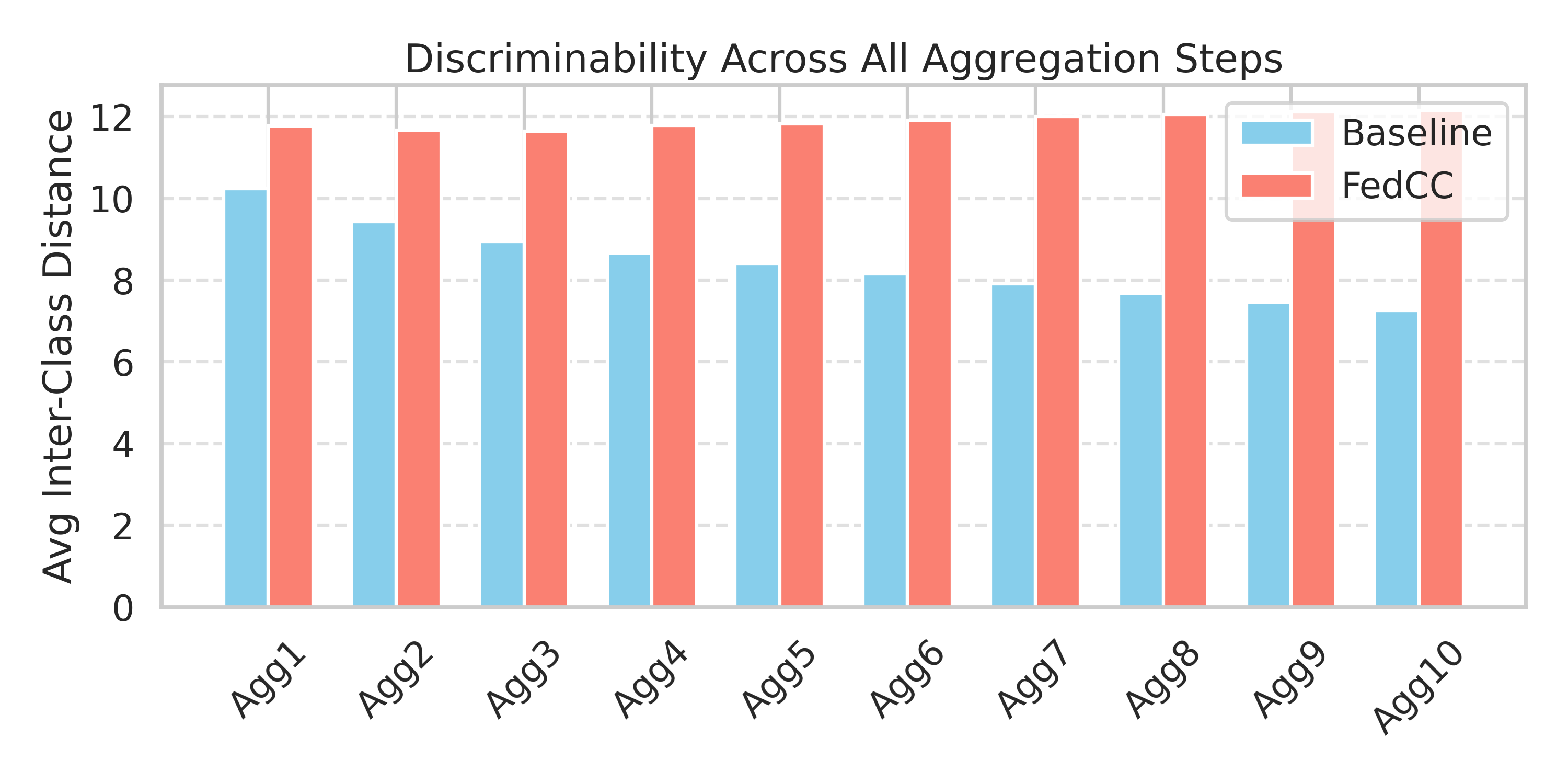}
    \end{minipage}
    \hfill
    \begin{minipage}{0.48\textwidth}
        \centering
        \includegraphics[width=\linewidth]{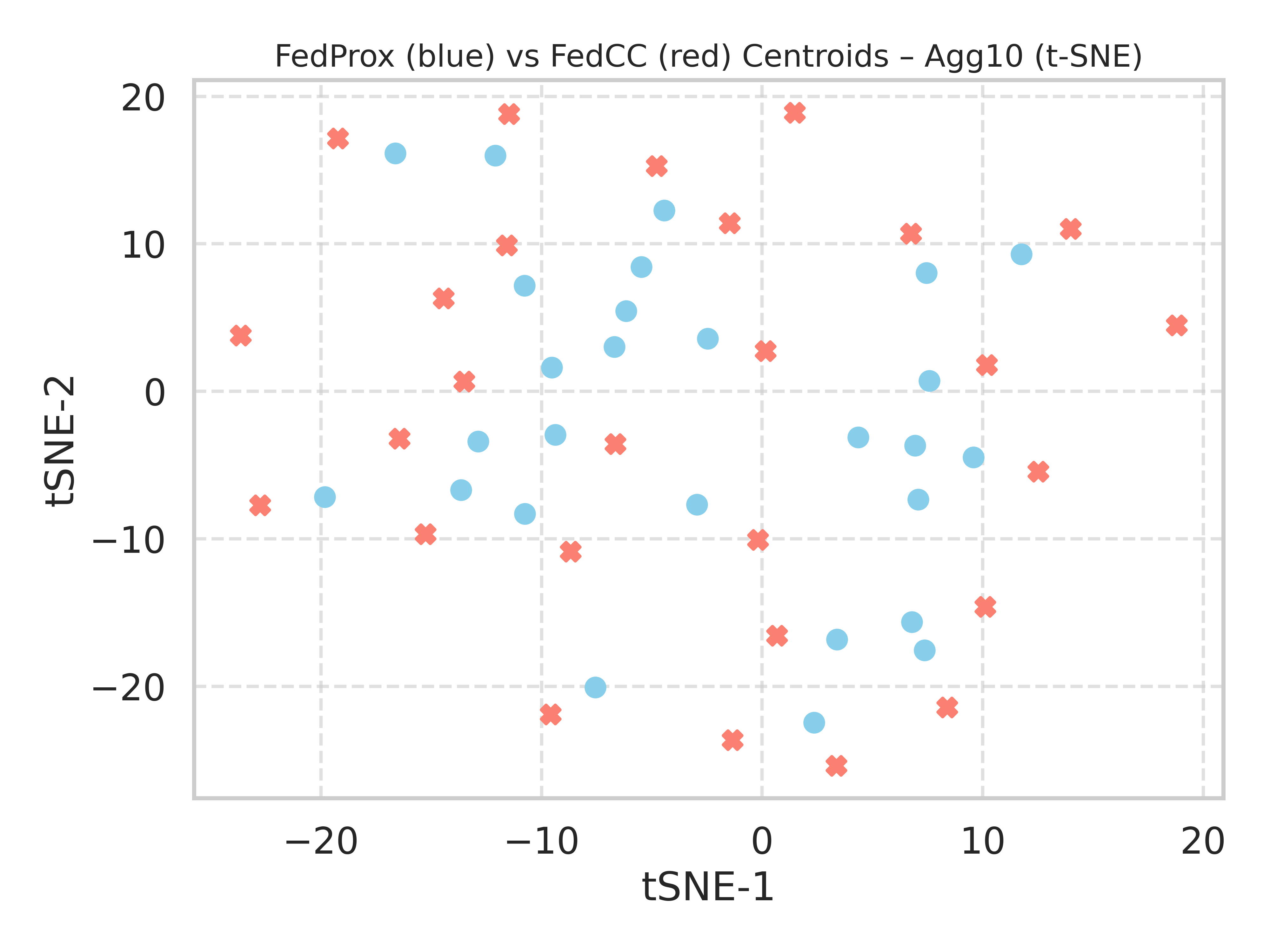}
    \end{minipage}
    \caption{Inter-class discriminability comparison for FedProx and FedProx+FedCC.}
    \label{fig:fedprox_qualitative}
\end{figure}

\begin{figure}[!h]
    \centering
    \begin{minipage}{0.48\textwidth}
        \centering
        \includegraphics[width=\linewidth]{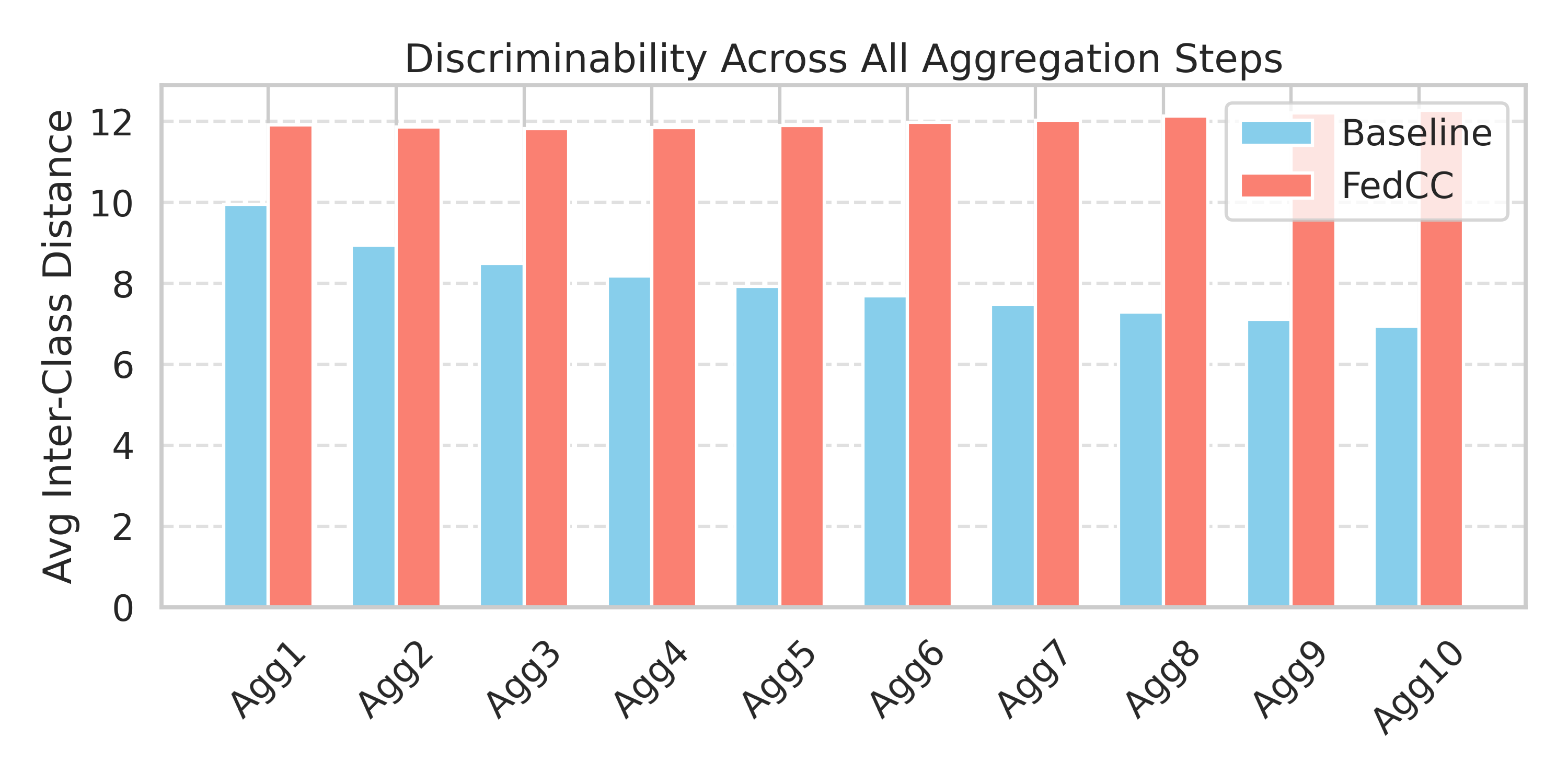}
    \end{minipage}
    \hfill
    \begin{minipage}{0.48\textwidth}
        \centering
        \includegraphics[width=\linewidth]{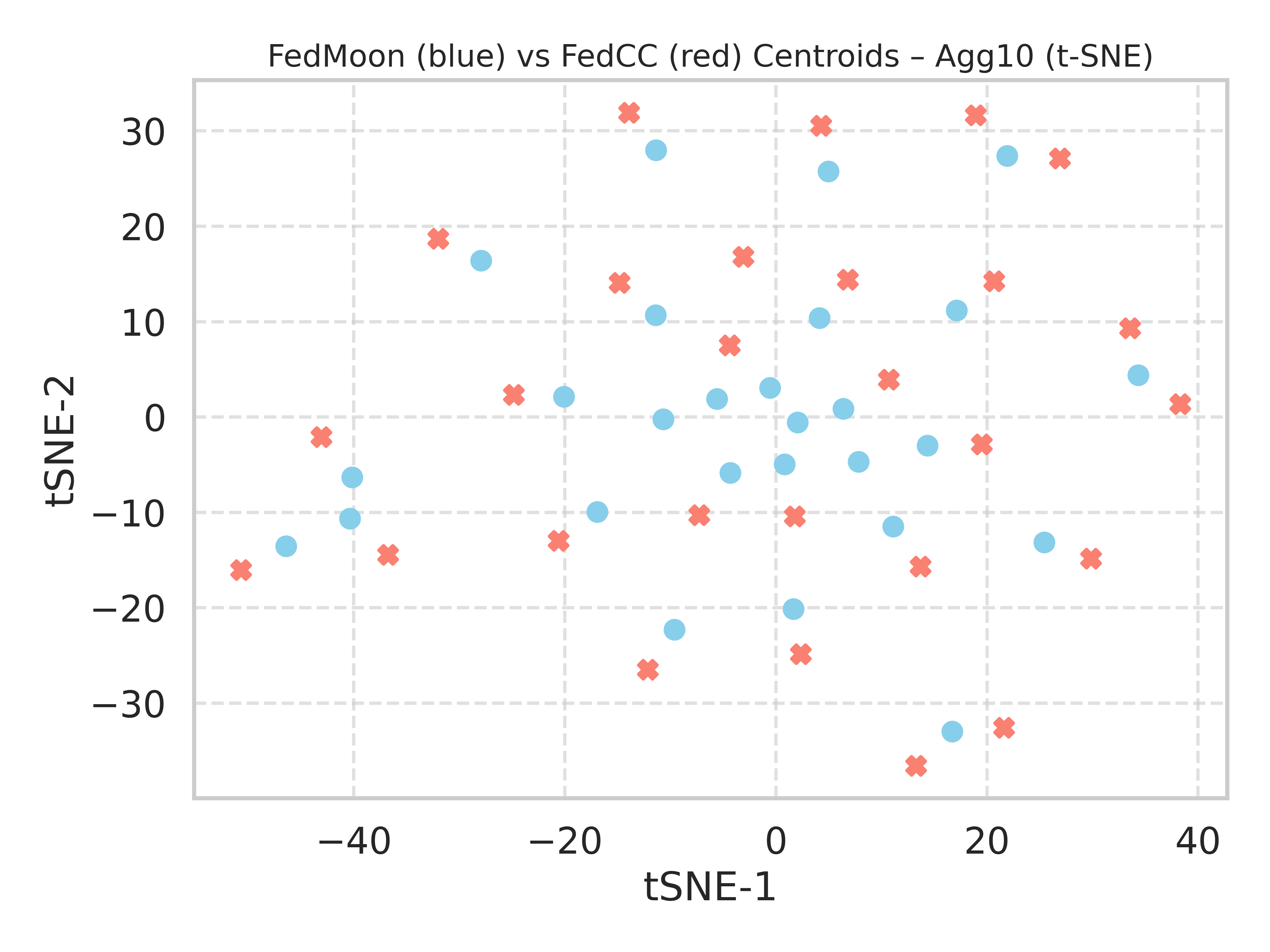}
    \end{minipage}
    \caption{Inter-class discriminability comparison for FedMoon and FedMoon+FedCC.}
    \label{fig:fedmoon_qualitative}
\end{figure}

We have included the t-SNE visualizations for FedProx vs. FedProx+FedCC (Figure~\ref{fig:fedprox_qualitative}) and FedMoon vs. FedMoon+FedCC (Figure~\ref{fig:fedmoon_qualitative}), when the encoder parameter aggregation is identical.

Similarity with Figure~\ref{fig:qualitative} highlights the fundamental mechanism of our framework: the improved separability is a direct result of the \texttt{FedCC} aggregation step, which explicitly enforces discriminability in the centroid space.

\end{document}